\documentclass{article}

\usepackage[main, preprint]{neurips_2026}

\usepackage[utf8]{inputenc} 
\usepackage[T1]{fontenc}    
\usepackage{hyperref}       
\usepackage{url}            
\usepackage{booktabs}       
\usepackage{amsfonts}       
\usepackage{nicefrac}       
\usepackage{microtype}      
\usepackage{xcolor}         

\usepackage{booktabs}

\usepackage{graphicx}
\usepackage{subcaption}
\usepackage{amsmath}
\usepackage{multirow}
\usepackage[table]{xcolor}
\usepackage{amssymb}

\usepackage{wrapfig}

\usepackage[normalem]{ulem}
\usepackage{float}

\title{RAwR: Role-Aware Rewiring via Approximate Equitable Partition}

\author{%
Riccardo~Porcedda\textsuperscript{1,2,*}
\And
Giuseppe~Squillace\textsuperscript{5,1,*}
\And
Bastian~Epping\textsuperscript{6}
\AND
Andrea~Vandin\textsuperscript{1,7}
\And
Michael~Schaub\textsuperscript{6}
\And
Mirco~Tribastone\textsuperscript{5}
\And
Francesca~Chiaromonte\textsuperscript{1,3,4}
\\[1.2ex]
\textsuperscript{1}Department of Excellence L'EMbeDS, Sant'Anna School of Advanced Studies, Pisa, Italy\\
\textsuperscript{2}Department of Computer Science, University of Pisa, Italy\\
\textsuperscript{3}Department of Statistics, The Pennsylvania State University, USA\\
\textsuperscript{4}Huck Institutes of the Life Sciences, The Pennsylvania State University, USA\\
\textsuperscript{5}IMT School for Advanced Studies, Lucca, Italy\\
\textsuperscript{6}Computational Network Science, RWTH Aachen University, Aachen, Germany\\
\textsuperscript{7}DTU Technical University of Denmark, Lyngby, Denmark\\[0.8ex]
\texttt{\{riccardo.porcedda,andrea.vandin\}@santannapisa.it}\\
\texttt{\{giuseppe.squillace,mirco.tribastone\}@imtlucca.it}\\
\texttt{\{epping,schaub\}@netsci.rwth-aachen.de}\\
\texttt{fxc11@psu.edu}\\[0.8ex]
\textsuperscript{*}Equal contribution.
}

\begin{document}

\maketitle

\begin{abstract}

    While Graph Neural Networks (GNNs) have demonstrated significant efficacy in node classification tasks, where predictions rely on local neighborhood information, the performance of GNNs often drops when prediction tasks depend on long-range interactions. 
    These limitations are attributed to phenomena such as oversquashing, where structural bottlenecks restrict signal propagation across the network topology. 
    To address this challenge, we introduce RAwR, a computationally efficient rewiring framework that augments the input graph with a quotient graph derived from
    equitable partitions. 
    This approach facilitates accelerated communication between nodes that share identical structural roles, as identified by the Weisfeiler-Leman graph coloring, and thereby reduces the total effective resistance of the system. 
    Furthermore, by employing an approximate definition of the equitable partition, RAwR enables a controllable reduction of the quotient graph, which, in its most condensed state, recovers the conventional Master Node rewiring technique. 
    Empirical evaluations across a diverse suite of benchmarks—including homophilic, heterophilic, and synthetic long-range datasets—demonstrate that RAwR achieves state-of-the-art results. 
    Our contribution is further supported by an analytical investigation using a teacher-student model of linear GNNs, which elucidates the theoretical foundations of role-based rewiring. 
    This analysis leads to the formulation of Spectral Role Lift (SRL), a metric designed to identify the optimal approximate equitable partition for maximizing predictive performance.
\end{abstract}

\section{Introduction}
\label{sec:intro}

Message-passing Graph Neural Networks (GNNs) learn node representations by repeatedly aggregating information along the edges of an input graph~\citep{scarselli2008graph,gilmer2017mpnn,kipf2017gcn,velickovic2018gat,xu2019gin}. This locality bias is effective when the relevant signal for prediction is contained in short-range neighborhoods. However, many node classification tasks depend on information that is structurally distant. In these settings, increasing the number of message-passing layers is not a complete solution: long-range signals may still be compressed through narrow topological bottlenecks before reaching the target node. This phenomenon, known as oversquashing, limits the ability of GNNs to exploit non-local dependencies~\citep{alon2020bottleneck,topping2022iclr,black2023icml}.

Graph rewiring addresses this limitation by modifying the topology on which message passing is performed. By adding edges or virtual nodes (VNs), rewiring methods can reduce structural bottlenecks without changing the downstream GNN architecture. Existing methods instantiate this idea in different ways: a simple and widely used VN-based instance is the Master Node construction, which adds a single global virtual node connected to all nodes~\citep{gilmer2017mpnn,southern2025vn,cai2023mpnnvn}; curvature-based approaches modify the graph according to discrete Ricci curvature or related flow procedures~\citep{topping2022iclr,nguyen2023revisiting}; spectral methods improve expansion properties~\citep{karhadkar2023fosr}; more recent approaches combine rewiring with feature denoising, community and feature similarity, or triangulation-based constructions~\citep{linkerhagnerjoint,rubio-madrigal2025gnns,attali2025dynamic}. These methods show that topology modification can substantially improve message passing. Nevertheless, they mostly focus on making propagation easier in a global sense. For node classification, however, \emph{we posit that not all long-range communication paths are equally useful}.

Edges encode observed interactions, but observed adjacency is not always the most informative communication pattern for prediction. In heterophilic graphs, for instance, immediate neighbors may belong to different classes, while nodes with similar labels or functions may be far apart. More generally, node classes can be reflected not only in local adjacency, but also in the structural role that a node occupies in the graph. Thus, the graph may still contain useful information even when its edges are not directly aligned with label similarity. A rewiring method should therefore not only reduce bottlenecks, but also route information through graph-induced channels that are likely to be relevant for the downstream task.

We propose to use structural roles as such a routing principle. Structural roles identify nodes that occupy similar positions with respect to the surrounding topology, independently of their graph distance. This notion has a long history in network analysis and has been shown to be useful for node classification and representation learning~\citep{lorrain1971structural,borgatti1992notions,rolx_2012,peel2014active,park_role_2020,cui_pos_struct_2022}. In this work, roles are obtained from equitable partitions. An equitable partition groups nodes that have the same number of neighbors in every block, inducing a quotient graph that exactly summarizes role-to-role interactions~\citep{godsil1997compact,everett1996exact}. This construction is closely related to color refinement and the one-dimensional Weisfeiler--Leman test: stable 1-WL color classes induce equitable partitions, and both mechanisms capture structural distinctions through iterative neighborhood-count profiles.
Approximate equitable partitions (AEPs) further relax exact role equivalence through a tolerance parameter, making the construction suitable for noisy real-world graphs~\citep{scholkemper2023neurips,squillace2024icdm,squillace2026tkde}.

Here we introduce RAwR, Role-Aware Rewiring, a computationally efficient and model-agnostic rewiring framework based on approximate equitable partitions. Given a tolerance $\varepsilon$, RAwR computes an $\varepsilon$-AEP of the input graph and assigns one virtual representative node to each role. The first variant of RAwR, which we call RepNodes, connects every node to its representative VN, creating two-hop communication paths between nodes that may be distant in the original topology but structurally similar. In a second variant, which we call RepEdges, representative VNs are also connected according to the unweighted quotient graph, enabling communication between interacting roles.

RAwR therefore preserves the original graph while adding a role-aware communication layer whose granularity is controlled by $\varepsilon$: fine partitions produce many role-specific representatives, whereas the fully collapsed partition recovers the Master Node as a limit case. Unlike rewiring methods that only improve global propagation, RAwR also determines where propagation should be facilitated, routing information through structural pathways induced by the quotient graph.

\begin{figure*}[t]
    \centering
    \includegraphics[width=\linewidth]{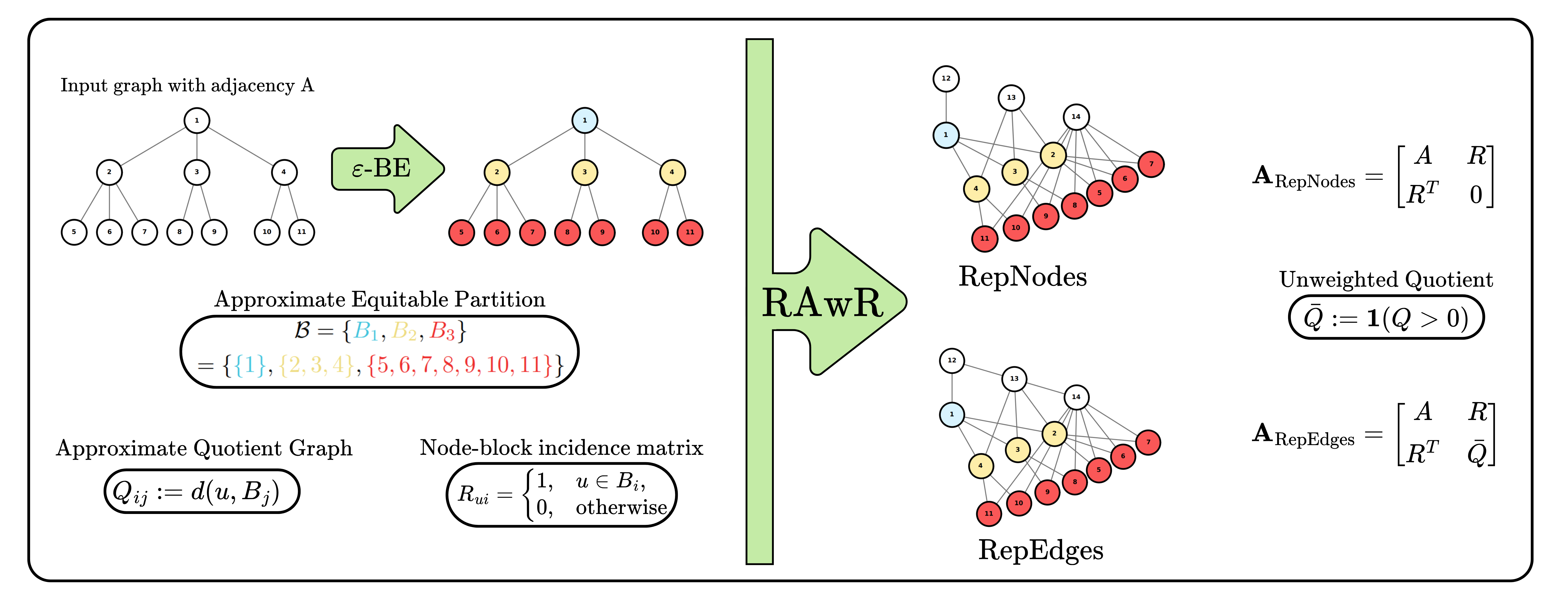}
    \caption{
    Overview of RAwR. Given an input graph and a tolerance $\varepsilon$, RAwR computes an $\varepsilon$-AEP with the $\varepsilon$-BE algorithm~\cite{squillace2024icdm,squillace2026tkde}, grouping nodes with similar structural roles. It then adds one virtual node per role. RepNodes connects each node to its role representative, while RepEdges additionally connects representatives according to the unweighted quotient matrix $\bar Q=\mathbf{1}(Q>0)$. For $\varepsilon=0$ the partition is exact; larger $\varepsilon$ values merge increasingly similar roles.}
    \label{fig:RAwR-overview}
\end{figure*}

\paragraph{Contribution.}
We summarize our contribution as follows. First, we introduce RAwR\footnote{https://zenodo.org/records/20054428}, a computationally efficient and model-agnostic graph rewiring method that augments the input topology with a quotient graph derived from an approximate equitable partition. While existing rewiring methods mainly focus on alleviating oversquashing by improving global signal propagation, RAwR also addresses where information should be routed, adding communication paths between nodes that are topologically distant but structurally similar.

Second, we provide structural evidence that RAwR alleviates long-range propagation bottlenecks by reducing the mean effective resistance of the input graph.

Third, we perform an extensive empirical evaluation across homophilic, heterophilic, and synthetic long-range benchmarks. For node classification, RAwR obtains the best average rank among the compared rewiring methods. We further evaluate against GRAIN, a recent model designed to address oversquashing, and show that RAwR is not only competitive with architecture-level solutions, but also complementary to them, since applying RAwR on top of GRAIN further improves performance.

Finally, we support the proposed framework through a teacher-student analysis of linear GNNs. This formalizes how role-aware rewiring modifies the spectral response of message-passing models and explains when the original graph is insufficient to recover the correct node labels. A byproduct of this analysis is Spectral Role Lift (SRL), a measure that provides a criterion for selecting the best approximate equitable partition.

\section{Background}
\label{sec:notation}

Let $G=(V,E)$ be a finite, undirected graph with $n = |V|$ nodes and adjacency
matrix $A \in \{0,1\}^{n \times n}$.

\paragraph{Equitable Partition (EP).}
A node partition is a set $\mathcal{B}=\{B_1,\dots,B_k\}$ of non-empty pairwise disjoint
blocks whose union is $V$.\\
For $u \in V$ and $B_j\in \mathcal{B}$, define the block-degree $
d(u,B_j) := \bigl|\{v \in B_j : (u,v)\in E\}\bigr|$.
The block-degree vector of $u$ is 
$d_{\mathcal{B}}(u) := \bigl(d(u,B_1),\dots,d(u,B_k)\bigr) \in \mathbb{N}^k$.
The partition $\mathcal{B}$ is equitable if for every $i \in \{1,\dots,k\}$ and every $u,v \in B_i$, $d_\mathcal{B}(u) = d_\mathcal{B}(v)$.

Let $R \in \mathbb{R}^{n \times k}$ be the node-block incidence matrix $R_{ui} = \mathbf{1}\{u \in B_i\}$.
For an equitable partition, the quotient matrix $Q \in \mathbb{N}^{k \times k}$ is defined by
$Q_{ij} := d(u,B_j)$ for any $u \in B_i$, and satisfies the matrix relation
\begin{equation}
AR = RQ.
\label{eq:AR_RQ}
\end{equation}

\paragraph{Approximate equitable partition (AEP).}
Fix a tolerance $\varepsilon \ge 0$ and a partition $\mathcal{B}=\{B_1,\dots,B_k\}$.
We say that $\mathcal{B}$ is an approximate equitable partition with tolerance $\varepsilon$  (in short, $\varepsilon$-AEP) if for every $i \in \{1,\dots,k\}$ and every $u,v \in B_i$,
\begin{equation}
\|d_\mathcal{B}(u) - d_\mathcal{B}(v)\|_\infty \leq \varepsilon,
\label{eq:eps_equitable}
\end{equation}
with $\|\cdot\|_\infty$ being the infinity norm.

We highlight that, while we can still define $R$ and $Q$ as above, Eq.~\eqref{eq:AR_RQ} no longer holds.
In fact, we consider the following relaxed version:  
\begin{equation} \label{eq: AR_RQ_eps}
    ||AR-RQ||_\infty \leq \varepsilon.
\end{equation}

We illustrate in appendix \ref{app:extended_background} the resulting exact and approximate equitable partition for a simple running example.

\section{Role-Aware Rewiring}

Given a graph with adjacency matrix $A$, an $\varepsilon$-AEP with indicator matrix $R$ and a quotient graph $Q$ so that Eq.~\eqref{eq: AR_RQ_eps} holds, 
the RAwR-augmented graph has 
adjacency matrix
\begin{align}
	A_{\mathrm{RAwR}}
	&=
	\begin{bmatrix}
		A & R\\
		R^\top & Q
	\end{bmatrix}.
	\label{eq:ARAwR_exact}
\end{align}

The $\varepsilon$-AEP is efficiently obtained in polynomial time in the number of nodes using the partition-refinement algorithm $\varepsilon$-BE proposed in \cite{squillace2024icdm, squillace2026tkde}.
Furthermore, the tolerance parameter $\varepsilon$ allows to tune the number $k$ of VNs, and, for $\varepsilon > 0$ we generally get $k \ll n$, therefore time and space overhead for downstream GNN training remains subquadratic in the number of nodes.
It emerges that finding the \emph{right} value for the tolerance $\varepsilon$ is crucial. This will be treated in Section \ref{sec:roles_important}.

As it is often desirable to stay in the framework of undirected and unweighted graphs, but $Q$ is in general a weighted quotient graph, we also introduce two simplifications called \emph{RepNodes} and \emph{RepEdges}, where we set $Q=0$ or we use an unweighted quotient graph which we denote $\bar Q$.
The respective adjacency matrix are the following:

\begin{equation*}
A_{RepNodes} =
\begin{bmatrix}
	A & R\\
	R^T & 0
\end{bmatrix}
\hspace{2cm}
A_{RepEdges} =
\begin{bmatrix}
	A & R\\
	R^T & \bar Q
\end{bmatrix}
\end{equation*}

Specifically, with RepNodes, we build the augmented adjacency matrix
RepNodes ensures any two nodes in the same role communicate in at most two hops, since information that is “role-relevant” can propagate globally in 
	$O(1)$ message passing steps).
In the top-right of Figure \ref{fig:RAwR-overview}, a simple example of RepNodes is shown: we can see that one node per role has been added (\texttt{12}, \texttt{13}, \texttt{14}), each connected with the nodes of the corresponding role. 
With RepEdges, in addition to RepNodes we connect representative nodes according to the quotient interactions.
Roughly speaking, this graph represents how the different roles interact with each other within the system. 
%
In the bottom-right of Figure \ref{fig:RAwR-overview}, a simple example of RepEdges is shown: we can see that, in addition to  RepNodes, 
we also connect the three representative nodes. 

The node feature matrix $X_\mathrm{RAwR}$ (it is the same for both variants) needed for the GNNs is set as
$	X_\mathrm{RAwR} =
	\begin{bmatrix}
		X & 0\\
		0 & I_k
	\end{bmatrix}$,
where $X$ is node-feature matrix of the original graph and $I_k$ is the identity matrix (VNs are basically one-hot encoded).

RAwR requires no upstream training stage and produces a single rewired graph on which downstream tasks can be performed.

\section{Quantifying the impact of RAwR}
\label{sec:roles_important}

This section formalizes why and how role-based virtual nodes can improve node classification accuracies.
To do this, we make the assumption that the observed graph is actually an incomplete instance of a richer latent structure, and node labels are generated by a teacher GNN on this latent graph.
Moreover, we assume that RAwR provides a proxy of this latent structure.
For simplicity we will focus on a linear filter GNN as the teacher model.

\subsection{Motivation}
%
	This analysis is useful to better understand how RAwR works. Furthermore, it allows to provide formal bounds on the error that a student GNN will make on the observed graph if the labels are generated according to our assumption, namely from a linear teacher GNN on the latent structure provided by RAwR. These bounds will generally depend on the training of the GNN, but, from these, we are able to extract a new metric called SRL (Spectral Role Lift). Notably, SRL is computed solely on the input rewired graph and correlates with node classification improvements, and it can be used as a proxy for the selection of the tolerance parameter $\varepsilon$. 
	Indeed, while this formal derivation holds for $\varepsilon=0$, i.e., for RaWR with exact EP, in Section~\ref{sec:exps} we study its practical validity also for non-zero $\varepsilon$.
%
The reasoning for this is as follows:
if the SRL is small, this provides a small bound on the error a student GNN does on the given data, thus we may conclude that
GNNs on observed and rewired graphs will obtain similar performances.
	However, if the SRL is large, this means that the student GNN poorly approximates the labels from the teacher model, thus we expect a performance improvement by the GNN trained on the RAwR graph.

\subsection{Formal derivation of SRL by teacher-student model}
\label{sec:ts_model}

\paragraph{Teacher-student model.}
Let us now consider the rewired graph from Eq.~\eqref{eq:ARAwR_exact}, and let $S_{\mathrm{obs}}$ and $S_{\mathrm{RAwR}}$ denote the corresponding normalized shifts (normalized adjacency with self-loops) for adjacency matrices $A$ and $A_{\mathrm{RAwR}}$ respectively.
According to our teacher-student model, we assume the true labels $y_\text{true}\in\mathbb{R}^{n\times d_\text{out}}$ to be generated as
$
	\label{eq:teacher_definition}
	y_\text{true} = \Big(h(S_\mathrm{RAwR}) X_\mathrm{RAwR} \prod_{l=1}^L W_t^{(l)}\Big)_{1:n,:}
$
where $h$ is the polynomial filter implemented by the GNN, 
$W_t^{(l)}$ are some fixed weight matrices and $d_\text{out}$ is the number of label classes.

We note that we use python-style indexing to solely consider the output of the GNN on the real nodes, omitting outputs on the virtual nodes.

For the output of the student on the original graph we assume
$
	\label{eq:student_definition}
	y_\text{obs} = h(S_\mathrm{obs}) X \prod_{l=1}^L W_s^{(l)}
$
where the shift operator and the weight matrix differs from the teacher.
The student is trained on the mean-squared error
$
	\mathrm{MSE}(W_s) = \frac{||y_\text{true} -  y_\text{obs}(W_s)||^2}{n d_{\text{out}}}
$
where we made the dependence on the student weights explicit with $W_s$ the set of all $W_s^{(l)}$.
We now 
sketch on how to derive the SRL metric as an upper bound on the student error while the full calculation is provided in Appendix \ref{app:roles_important}.

\paragraph{Bound on the student's performance.}
We start by noticing that linear GNNs trained with gradient descent always converge to their global optimum, the weights of which we will call $W^\star_s$ \cite{patel2025convergencegradientbasedtraining}.
Thus we have
\begin{align}
	\mathrm{MSE}(W_s) \leq \mathrm{MSE}(W_t) = \sum_c \frac{||e^{(c)}||^2}{n d_{\text{out}}}
\end{align}
where we defined $e^{(c)} = \big( y_\text{true} -  y_\text{obs}(W_t) \big)_{c,:} \in \mathbb{R}^{n}$ the difference of outputs between teacher and student for a specific output dimension $c$.
	Then we have
	\begin{align}
		\label{eq:srl_err_lin_relation}
		\sum_c ||e^{(c)}||^2 \leq \kappa_0 + \kappa_{\text{max}} E_{tot} \mathrm{SRL}
	\end{align}
	with constants $\kappa_0$ and $\kappa_{\text{max}}$ which depend on the trained GNN, $E_{tot}$ and SRL only on the input rewired graph, with
    $
    		\mathrm{SRL} = \rho \sum_{j=1}^k \omega_j\,\Delta_j^2.
    $
	Intuitively, $\rho$ measures the energy of labels in the role subspace, $\Delta_j$ is the difference of role-relevant eigenvalues in $S_{\mathrm{obs}}$ and $S_{\mathrm{RAwR}}$, while $\omega_j$ weights the contribution of each role. 
	For a complete derivation and all definitions we provide each step and quantity in Appendix \ref{app:roles_important}.

	\subsection{Validation of SRL via simulated data}
    \begin{wrapfigure}[14]{r}{0.4\linewidth}
        \vspace{-1.9cm}
		\centering
		\includegraphics[width=0.4\columnwidth]{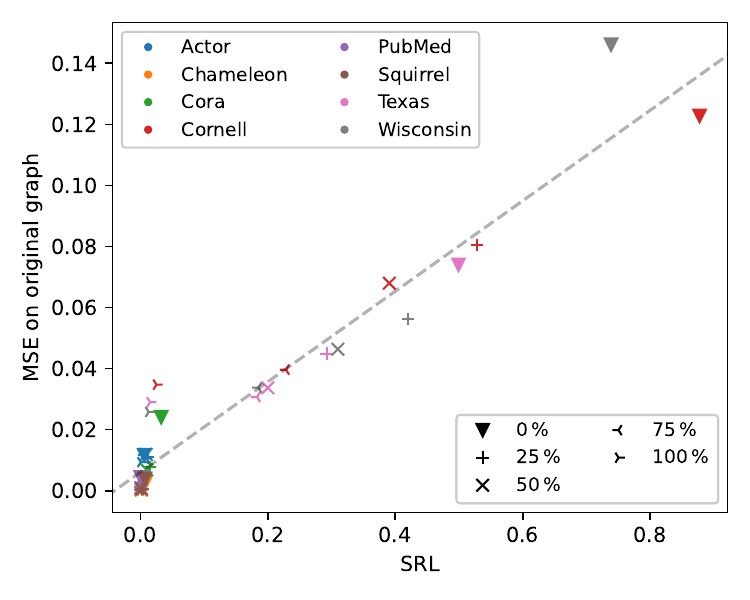}
        \vspace{-0.75cm}
		\caption{
        Scatter plot of SRL and mean squared error (MSE) in the teacher-student setup from Section \ref{sec:ts_model} applied on real world datasets.
			The $\varepsilon$, denoted by different symbols, are chosen as degree sequence percentiles (in the legend).
			The dashed line shows a linear fit.
			Details on experimental setup are in Appendix \ref{app:details_ts_experiment}.
            }
		\label{fig:ts_experiment}
    \end{wrapfigure}
	In Figure \ref{fig:ts_experiment}, we validate our approach in the teacher-student setup numerically. 
	We train the student model on the original graph with labels generated by a teacher on the rewired graph.
	As expected, the $\mathrm{SRL}$ metric provides a good measure when the student model is able to learn the presented task and when it fails.
	Specifically, a large SRL implies that GNN functions implemented on the rewired graph cannot be approximated well by GNNs on the original graph.
	This results in a larger error after training.
	We added a fit on the linear relation between SRL and error predicted by Eq.~\eqref{eq:srl_err_lin_relation} and see a good agreement with the observed SRL values and train error.
	
	\subsection{Further considerations on SRL as an heuristic} \label{sec:srl_heuristic}
	We highlight that SRL is obtained by assuming linear GNNs that act as a polynomial filter: while a GCN could be linearized to match this assumption, the same task is not trivial for GAT and GIN. For this reason, we expect SRL to be more meaningful for  GCNs than for the other architectures.
	
	Furthermore, the link between SRL and RAwR improvements has been shown by starting from an (exact) EP for ease of mathematical derivation. When the value of $\varepsilon$ increases, the quotient graph rapidly reduces to fewer nodes, and it is therefore unavoidable for the role subspace energy $\rho$ to collapse. While labels may not live anymore in the role subspace, the virtual nodes can still help the message passing along nodes with same labels that are normally far away in the graph.
	To account for this, we add to our framewrok the 2NCS metric from Cavallo et al. \cite{cavallo20222hopneighborclasssimilarity}. For doing this, we  define $\text{SRL}^*$ as:
	\begin{equation}
		\text{SRL}^* = \sqrt[4]{\rho} \, z(\text{SRL}) + (1-\sqrt[4]{\rho}) \, z(\text{2NCS})
	\end{equation}
	being $z(x)$ the z-score of $\sqrt x$ (we apply the square root transformation to reduce the skewness of the metrics and then we consider the z-scores in order to have a common scale for comparison).
	%
	%
    In Appendix \ref{app:srl_corr} we experimentally study the correlation between $\text{SRL}^*$ and mean test accuracies.

\section{Experiments}
\label{sec:exps}

\subsection{Setup}
\label{sec:setup}
We conduct a systematic evaluation on nine widely used datasets that jointly cover homophilous and heterophilous regimes: Actor, Chameleon, Citeseer, Cora, Cornell, PubMed, Squirrel, Texas, and Wisconsin.
These datasets are widely adopted to evaluate GNNs (e.g., in \cite{GeomGCN}), as well as rewiring methods (\cite{nguyen2023revisiting,topping2022iclr, attali2024delaunay, linkerhagnerjoint, rubio-madrigal2025gnns, attali2025dynamic}).

We also believe that better assessment is required to fairly evaluate rewiring methods efficacy on long-range graphs. Prior works referred to the datasets \emph{PascalVOC-SP} and \emph{COCO-SP } (\cite{dwivedi2022long}) for node classification, but the actual validity of these benchmarks has been debated (\cite{bechler-speicher2025position}).
A recent publication (\cite{liang2026towards}) proposed \emph{City-Networks}, a long-range graph benchmark built from four large-scale city maps where node labels are obtained by binning the node eccentricity values.
Nonetheless, the size of these graphs creates a digital divide that is addressed by \cite{miglior2026can} who propose the \emph{ECHO} dataset, an ensemble of smaller synthetic graphs specifically designed for long-range propagation.
This last one is the additional benchmark adopted in this work.
All graphs are treated as undirected and unweighted.

For each dataset, we train GCN first on the original graph (i.e., the baseline) and then on graphs augmented with RAwR, considering both RepNodes and RepEdges. We considered other architectures like GIN and GAT and we report the extensive analysis in Appendix \ref{app:additional_experiments}.

For each graph, we choose five values of $\varepsilon$ from the degree sequence percentiles -- 
\{$0, 0.25, 0.5, 0.75, 1$\} -- and we repeat the experiments for each $\varepsilon$-AEP partition obtained. By construction, the $0^{\circ}$ percentile recovers the exact equitable partition, whereas the $100^{\circ}$ percentile collapses the partition to a single block, coinciding with the Master Node.
On these configurations, we both execute a grid search on the validation set to find the $\varepsilon$ which obtains the best accuracy and also choose it by using the $\text{SRL}^*$ heuristic described in Section \ref{sec:srl_heuristic} using only node labels from the training set.

We compare RAwR to state-of-the-art rewiring techniques already mentioned in the Related Work section \ref{sec:intro}: BORF, FOSR, SDRF, JDR, ComFy and TRIGON.

\paragraph{Random partitions.}
As a control, we replace the $\varepsilon$-AEP with random partitions matched in the number of blocks, and compare them with RepNodes without node features, so as to isolate the effect of the topology. We run this control in the small-$\varepsilon$ regime, where partitions are fine-grained and RAwR is most sensitive to the role assignment. For each $\varepsilon$, we sample five random assignments and report mean test accuracy with standard error.

\paragraph{Complementarity to SOTA model architectures.} While a direct comparison of model architectures against RaWR, being a graph rewiring method, is not a fair comparison as any GNN architecture is suitable to work on the rewired graph (including the ones we compare against here), we show in the appendix the following results (Table \ref{tab:gcn_grain_rawr_comparison}): GCN+RAwR beats state-of-the-art architectures designed to address oversquashing; RAwR, being model-agnostic, is capable of further improving those state-of-the-art models, showing its complementarity.

\subsection{Main results}
\label{sec:results}

We show in this section how our experimental setup provides evidence that RAwR effectively addresses the problem of oversquashing and that it achieves state-of-the-art results on the task of node classification. We focues here on two-layer GCN models. A more extensive evaluation across additional depths, backbones, and configurations is provided in Appendix~\ref{app:details_ts_experiment}.

\begin{figure}[t]
\centering
\begin{subfigure}{.49\columnwidth}
	\centering
 \includegraphics[width=\textwidth]{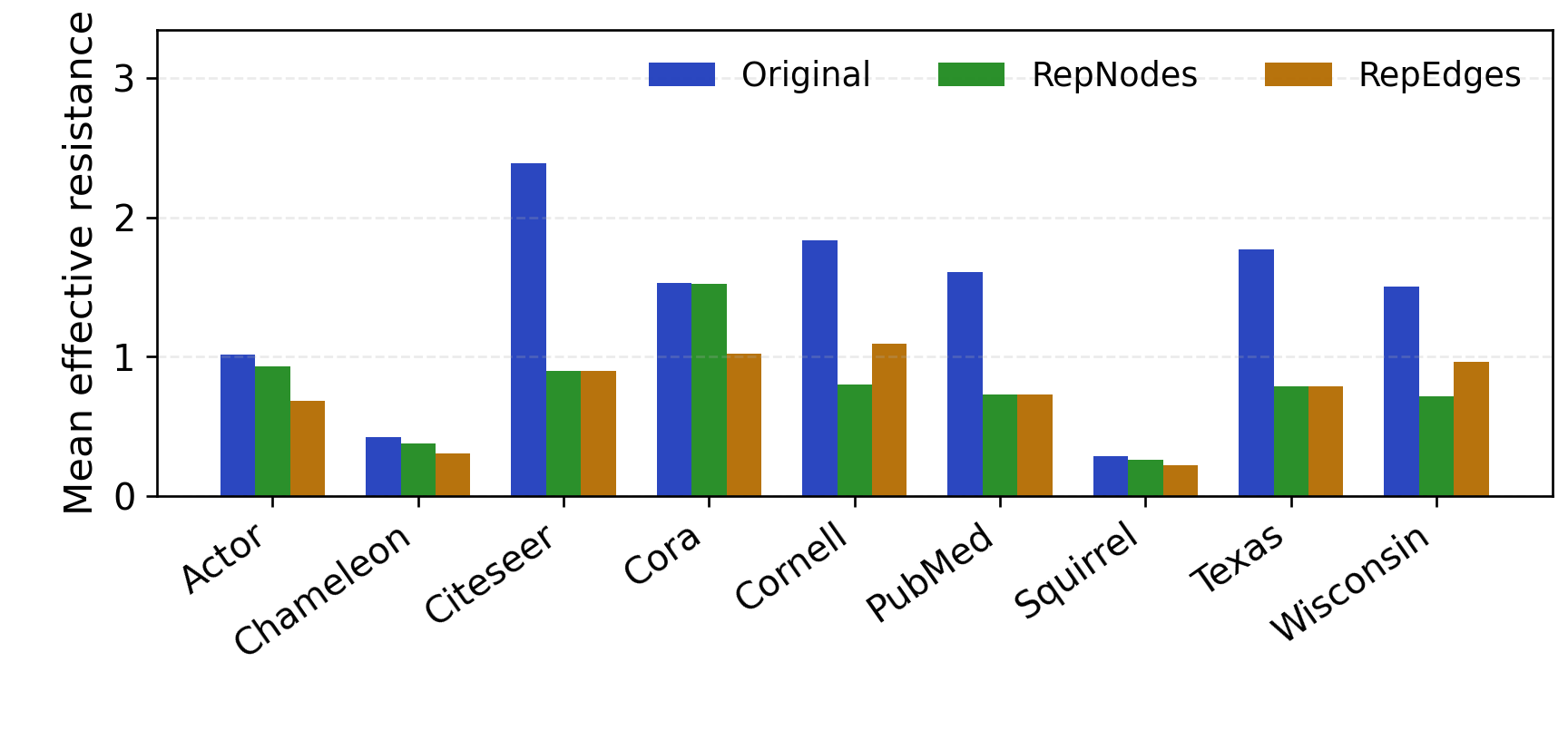}
	\caption{}
	\label{fig:exact}
\end{subfigure}
\begin{subfigure}{.49\columnwidth}
        \centering
 \includegraphics[width=\textwidth]{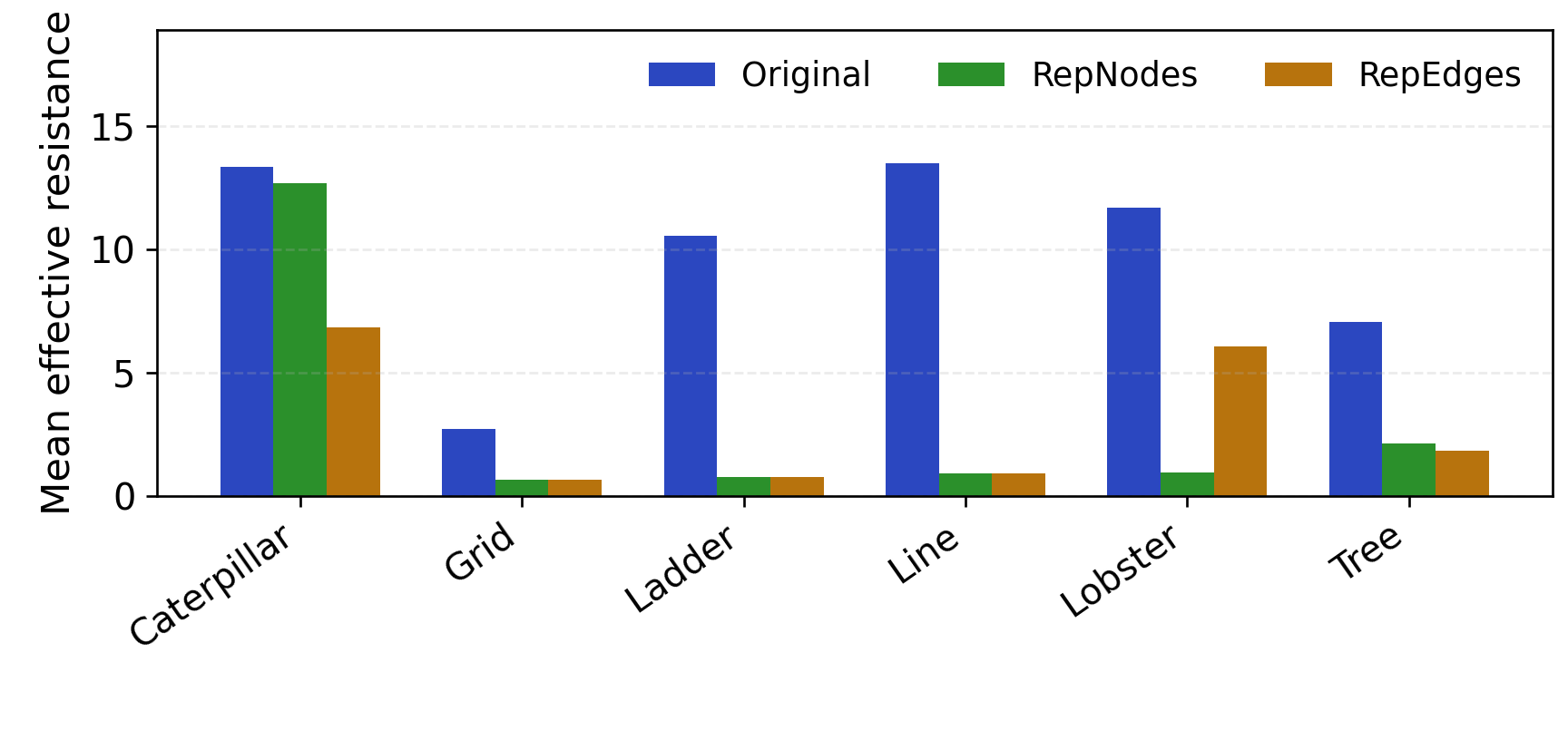}
	\caption{}
	\label{fig:approx}
\end{subfigure}
\caption{Mean effective resistance of original graphs and RAwR-rewired graph for (a) homophilic and heterophilic datasets (b) and the \emph{ECHO benchmark}.
}
\label{fig:effRes}
\end{figure}

\begin{table}[t]
\centering
\resizebox{\textwidth}{!}{%
\begin{tabular}{lccccccccc c}
\toprule
 & Actor & Chameleon & Citeseer & Cora & Cornell & PubMed & Squirrel & Texas & Wisconsin & Avg.Rank \\
\cmidrule{1-10} \cmidrule(l){11-11}
\multicolumn{11}{c}{\emph{WITH NODE FEATURES}}
\\
\cmidrule{1-10} \cmidrule(l){11-11}
Baseline & 25.00 $\pm$ 0.74 & 65.02 $\pm$ 0.59 & 75.18 $\pm$ 0.59 & 82.00 $\pm$ 0.97 & 44.44 $\pm$ 0.00 & 86.15 $\pm$ 0.04 & 61.65 $\pm$ 0.22 & 33.33 $\pm$ 0.00 & 52.00 $\pm$ 0.00 & 7.44 \\
 BORF & 25.63 $\pm$ 0.23 & 66.43 $\pm$ 0.70 & 68.92 $\pm$ 0.35 & 82.96 $\pm$ 0.00 & 45.56 $\pm$ 4.16 & OOM & OOM & 36.67 $\pm$ 6.67 & 56.80 $\pm$ 2.99 & 7.22 \\
 FOSR & 25.50 $\pm$ 0.38 & 65.99 $\pm$ 0.33 & 68.49 $\pm$ 0.24 & 83.11 $\pm$ 0.18 & 44.44 $\pm$ 0.00 & 82.09 $\pm$ 0.06 & 61.77 $\pm$ 0.54 & 33.33 $\pm$ 0.00 & 55.20 $\pm$ 5.88 & 7.22 \\
SDRF & 25.37 $\pm$ 0.44 & 64.85 $\pm$ 0.58 & 69.04 $\pm$ 0.23 & 83.48 $\pm$ 0.18 & 44.44 $\pm$ 0.00 & OOM & 61.65 $\pm$ 0.31 & 33.33 $\pm$ 0.00 & 51.20 $\pm$ 1.60 & 8.00 \\
 JDR & 29.16 $\pm$ 0.11 & 66.52 $\pm$ 1.53 & 74.52 $\pm$ 0.53 & \underline{83.70 $\pm$ 0.41} & 41.11 $\pm$ 2.72 & \underline{87.02 $\pm$ 0.09} & 53.19 $\pm$ 0.67 & \underline{46.67 $\pm$ 2.72} & \underline{69.60 $\pm$ 1.96} & 4.78 \\
 ComFy & \underline{29.37 $\pm$ 0.75} & 64.32 $\pm$ 0.54 & \textbf{78.19 $\pm$ 0.50} & \textbf{87.26 $\pm$ 0.77} & 31.11 $\pm$ 3.04 & \textbf{87.95 $\pm$ 0.08} & 42.04 $\pm$ 0.52 & 30.00 $\pm$ 4.97 & 53.60 $\pm$ 2.19 & 6.33 \\
 TRIGON & \textbf{36.18 $\pm$ 1.87} & 29.43 $\pm$ 5.19 & 71.57 $\pm$ 2.05 & 78.00 $\pm$ 6.07 & \textbf{71.11 $\pm$ 2.48} & 80.73 $\pm$ 0.18 & 41.85 $\pm$ 1.16 & \textbf{70.00 $\pm$ 6.33} & \textbf{76.80 $\pm$ 3.35} & 6.22 \\
 MN & 25.16 $\pm$ 0.68 & 66.30 $\pm$ 0.84 & OOM & 81.70 $\pm$ 0.40 & 47.78 $\pm$ 2.87 & OOM & 61.17 $\pm$ 0.29 & 34.44 $\pm$ 2.34 & 48.80 $\pm$ 1.69 & 9.00 \vspace{0.15 cm}\\
 \emph{RepNodes (SRL*)} & 25.95 $\pm$ 0.37 & \textbf{73.83 $\pm$ 0.50} & 74.58 $\pm$ 0.74 & 81.56 $\pm$ 0.41 & \underline{56.67 $\pm$ 2.48} & 86.34 $\pm$ 0.20 & \textbf{65.77 $\pm$ 0.38} & 33.33 $\pm$ 0.00 & 56.00 $\pm$ 4.00 & \underline{4.33} \\
 \emph{RepNodes (best)} & - & - & \underline{75.24 $\pm$ 0.37} & 81.93 $\pm$ 0.66 & - & 86.46 $\pm$ 0.22 & - & 43.33 $\pm$ 2.48 & - & \textbf{3.00} \\
 \emph{RepEdges (SRL*)} & 25.50 $\pm$ 0.24 & \underline{71.37 $\pm$ 0.31} & 74.58 $\pm$ 0.74 & 81.48 $\pm$ 0.45 & 50.00 $\pm$ 0.00 & 86.34 $\pm$ 0.20 & \underline{63.46 $\pm$ 0.19} & 33.33 $\pm$ 0.00 & 51.20 $\pm$ 3.35 & 6.11 \\
 \emph{RepEdges (best)} & 25.68 $\pm$ 0.14 & - & \underline{75.24 $\pm$ 0.37} & 81.93 $\pm$ 0.66 & - & 86.46 $\pm$ 0.22 & - & 38.89 $\pm$ 5.56 & - & 4.56 \\
\cmidrule{1-10} \cmidrule(l){11-11}
\multicolumn{11}{c}{\emph{WITHOUT NODE FEATURES}}
\\
\cmidrule{1-10} \cmidrule(l){11-11}
 BORF & 25.50 $\pm$ 0.92 & 68.46 $\pm$ 0.72 & 68.55 $\pm$ 0.24 & 83.19 $\pm$ 0.18 & 47.78 $\pm$ 2.72 & OOM & OOM & \underline{40.00 $\pm$ 6.48} & \textbf{62.40 $\pm$ 3.20} & 6.67 \\
 FOSR & 25.82 $\pm$ 0.99 & 68.90 $\pm$ 0.72 & 68.31 $\pm$ 0.35 & 83.26 $\pm$ 0.15 & 44.44 $\pm$ 0.00 & 82.09 $\pm$ 0.06 & 61.19 $\pm$ 0.33 & 33.33 $\pm$ 0.00 & \underline{58.40 $\pm$ 1.96} & 5.78 \\
 SDRF & 25.82 $\pm$ 0.97 & 67.93 $\pm$ 0.85 & 68.67 $\pm$ 0.33 & 83.33 $\pm$ 0.23 & 44.44 $\pm$ 0.00 & OOM & 61.31 $\pm$ 0.37 & 34.44 $\pm$ 2.22 & 52.00 $\pm$ 2.53 & 6.33 \\
 JDR & \underline{25.95 $\pm$ 0.45} & 60.18 $\pm$ 0.77 & 69.46 $\pm$ 0.45 & 82.67 $\pm$ 0.28 & 44.44 $\pm$ 0.00 & \textbf{82.56 $\pm$ 0.12} & 51.08 $\pm$ 0.52 & 34.44 $\pm$ 2.22 & 50.40 $\pm$ 1.96 & 6.89 \\
 ComFy & \textbf{28.32 $\pm$ 0.72} & 65.20 $\pm$ 0.82 & \textbf{72.53 $\pm$ 0.39} & \textbf{84.15 $\pm$ 0.17} & 28.89 $\pm$ 6.09 & \underline{82.15 $\pm$ 0.27} & 53.88 $\pm$ 0.76 & 27.78 $\pm$ 0.00 & 51.20 $\pm$ 4.38 & 6.00 \\
 TRIGON & \textbf{28.32 $\pm$ 0.31} & 49.87 $\pm$ 12.91 & 56.39 $\pm$ 4.98 & 68.37 $\pm$ 7.29 & 6.67 $\pm$ 4.65 & 54.28 $\pm$ 3.33 & 54.31 $\pm$ 0.66 & 27.78 $\pm$ 0.00 & 54.40 $\pm$ 6.69 & 8.89 \\
 MN & 25.39 $\pm$ 0.90 & 68.24 $\pm$ 0.35 & OOM & 83.24 $\pm$ 0.32 & 46.67 $\pm$ 2.79 & OOM & 60.73 $\pm$ 0.53 & 35.56 $\pm$ 2.79 & 49.60 $\pm$ 2.01 & 8.22 \vspace{0.15 cm}\\


 \emph{RepNodes (SRL*)} & 25.24 $\pm$ 0.68 & \textbf{75.07 $\pm$ 0.23} & \underline{70.84 $\pm$ 0.13} & 83.30 $\pm$ 0.27 & \textbf{55.56 $\pm$ 0.00} & 81.95 $\pm$ 0.19 & \textbf{65.42 $\pm$ 0.76} & 33.33 $\pm$ 0.00 & 54.40 $\pm$ 2.07 & \underline{4.22} \\
 \emph{RepNodes (best)} & 25.39 $\pm$ 1.08 & - & - & 83.33 $\pm$ 0.00 & - & - & - & \textbf{43.33 $\pm$ 2.34} & - & \textbf{3.00} \\
 \emph{RepEdges (SRL*)} & 25.11 $\pm$ 0.65 & \underline{71.72 $\pm$ 0.28} & 69.04 $\pm$ 0.13 & 83.41 $\pm$ 0.16 & \underline{50.00 $\pm$ 0.00} & 81.91 $\pm$ 0.30 & \underline{63.50 $\pm$ 0.72} & 33.33 $\pm$ 0.00 & 48.80 $\pm$ 1.69 & 6.33 \\
 \emph{RepEdges (best)} & 25.39 $\pm$ 1.08 & - & 69.79 $\pm$ 0.29 & - & - & 81.93 $\pm$ 0.23 & - & 36.67 $\pm$ 4.68 & 49.60 $\pm$ 3.37 & 4.89 \\
\bottomrule
\end{tabular}}
\caption{GCN node classification test accuracies on homophilic and heterophilic datasets. For RAwR, we report both the result for the epsilon chosen by a grid search on validation set and the one chosen by the SRL* heuristic; if they coincide we report '-'. OOM stands for "Out Of Memory". Best result per dataset is \textbf{bold}, second-best is \underline{underlined}. We also report the Average Rank (lower is better).}
\label{tab:ncq1}
\end{table}

\begin{table}[t]
\centering
\resizebox{\textwidth}{!}{%
\begin{tabular}{lcccccc c}
\cmidrule{1-7} \cmidrule(l){8-8}
 & Caterpillar & Grid & Ladder & Line & Lobster & Tree & Avg.Rank \\
\cmidrule{1-7} \cmidrule(l){8-8}
\multicolumn{8}{c}{\emph{WITH NODE FEATURES}}
\\
\cmidrule{1-7} \cmidrule(l){8-8}
Baseline & 41.18 $\pm$ 0.00 & 71.82 $\pm$ 2.59 & 60.00 $\pm$ 6.39 & \textbf{100.00 $\pm$ 0.00} & 78.33 $\pm$ 7.45 & \textbf{100.00 $\pm$ 0.00} & 5.17 \\
 BORF & 43.53 $\pm$ 3.22 & 71.82 $\pm$ 2.03 & 54.29 $\pm$ 3.91 & 55.00 $\pm$ 32.60 & 83.33 $\pm$ 0.00 & \textbf{100.00 $\pm$ 0.00} & 6.33 \\
 FOSR & 40.00 $\pm$ 2.63 & 62.73 $\pm$ 2.03 & 58.57 $\pm$ 5.98 & 50.00 $\pm$ 0.00 & 78.33 $\pm$ 7.45 & \underline{82.50 $\pm$ 6.85} & 8.83 \\
 SDRF & 42.35 $\pm$ 2.63 & \underline{74.09 $\pm$ 2.59} & \textbf{64.29 $\pm$ 5.05} & \textbf{100.00 $\pm$ 0.00} & 78.33 $\pm$ 7.45 & \textbf{100.00 $\pm$ 0.00} & 4.00 \\
 JDR & 11.76 $\pm$ 0.00 & 10.45 $\pm$ 3.69 & 37.14 $\pm$ 7.00 & 20.00 $\pm$ 10.00 & 50.00 $\pm$ 0.00 & 15.00 $\pm$ 14.58 & 11.83 \\
 ComFy & 51.76 $\pm$ 6.44 & 58.18 $\pm$ 6.93 & \underline{62.86 $\pm$ 9.31} & \underline{70.00 $\pm$ 11.18} & 83.33 $\pm$ 0.00 & 42.50 $\pm$ 25.92 & 6.67 \\
 TRIGON & 68.24 $\pm$ 3.22 & 41.82 $\pm$ 5.23 & 45.71 $\pm$ 11.95 & 20.00 $\pm$ 20.92 & \textbf{95.00 $\pm$ 4.56} & 20.00 $\pm$ 18.96 & 8.33 \\
 MN & 44.71 $\pm$ 3.22 & \textbf{76.36 $\pm$ 2.59} & \underline{62.86 $\pm$ 3.19} & \textbf{100.00 $\pm$ 0.00} & \underline{91.67 $\pm$ 0.00} & \textbf{100.00 $\pm$ 0.00} & 2.33 \vspace{0.15 cm}\\
 \emph{RepNodes (SRL*)} & \textbf{76.47 $\pm$ 0.00} & 57.27 $\pm$ 1.90 & 55.71 $\pm$ 9.31 & \textbf{100.00 $\pm$ 0.00} & 75.00 $\pm$ 0.00 & \textbf{100.00 $\pm$ 0.00} & 5.50 \\
 \emph{RepNodes (best)} & - & \textbf{76.36 $\pm$ 2.59} & \underline{62.86 $\pm$ 3.19} & - & \underline{91.67 $\pm$ 0.00} & - & \textbf{1.33} \\
 \emph{RepEdges (SRL*)} & \underline{70.59 $\pm$ 0.00} & 68.64 $\pm$ 1.90 & 57.14 $\pm$ 8.75 & \textbf{100.00 $\pm$ 0.00} & 83.33 $\pm$ 0.00 & \textbf{100.00 $\pm$ 0.00} & 4.17 \\
 \emph{RepEdges (best)} & - & \textbf{76.36 $\pm$ 2.59} & \underline{62.86 $\pm$ 3.19} & - & \underline{91.67 $\pm$ 0.00} & - & \underline{1.67} \\
\cmidrule{1-7} \cmidrule(l){8-8}
\multicolumn{8}{c}{\emph{WITHOUT NODE FEATURES}}
\\
\cmidrule{1-7} \cmidrule(l){8-8}
Baseline & 38.82 $\pm$ 3.22 & 72.73 $\pm$ 4.25 & \underline{62.86 $\pm$ 3.19} & \textbf{100.00 $\pm$ 0.00} & 76.67 $\pm$ 3.73 & \textbf{100.00 $\pm$ 0.00} & 4.17 \\
 BORF & 38.82 $\pm$ 5.26 & 72.27 $\pm$ 5.18 & 57.14 $\pm$ 5.05 & 85.00 $\pm$ 33.54 & \underline{85.00 $\pm$ 3.73} & \textbf{100.00 $\pm$ 0.00} & 6.50 \\
 FOSR & 36.47 $\pm$ 7.67 & 66.36 $\pm$ 1.02 & 55.71 $\pm$ 3.19 & 50.00 $\pm$ 17.68 & 73.33 $\pm$ 3.73 & \underline{87.50 $\pm$ 8.84} & 10.33 \\
 SDRF & 38.82 $\pm$ 3.22 & \underline{73.64 $\pm$ 2.03} & 58.57 $\pm$ 9.31 & \textbf{100.00 $\pm$ 0.00} & 76.67 $\pm$ 3.73 & \textbf{100.00 $\pm$ 0.00} & 5.17 \\
 JDR & 44.71 $\pm$ 2.88 & 60.00 $\pm$ 1.11 & 54.29 $\pm$ 7.28 & \underline{95.00 $\pm$ 10.00} & 73.33 $\pm$ 3.33 & \textbf{100.00 $\pm$ 0.00} & 8.33 \\
 ComFy & 64.71 $\pm$ 4.16 & 61.36 $\pm$ 3.94 & \underline{62.86 $\pm$ 5.98} & 75.00 $\pm$ 30.62 & 78.33 $\pm$ 4.56 & 50.00 $\pm$ 12.50 & 7.33 \\
 TRIGON & 67.06 $\pm$ 24.47 & 59.09 $\pm$ 12.96 & \textbf{65.71 $\pm$ 5.98} & 35.00 $\pm$ 22.36 & 20.00 $\pm$ 27.39 & 50.00 $\pm$ 0.00 & 8.83 \\
 MN & 45.88 $\pm$ 2.63 & \textbf{75.45 $\pm$ 1.90} & 61.43 $\pm$ 3.91 & \textbf{100.00 $\pm$ 0.00} & \textbf{91.67 $\pm$ 0.00} & \textbf{100.00 $\pm$ 0.00} & 2.50 \vspace{0.15 cm}\\
 \emph{RepNodes (SRL*)} & \textbf{76.47 $\pm$ 0.00} & 60.00 $\pm$ 2.03 & 60.00 $\pm$ 3.91 & \textbf{100.00 $\pm$ 0.00} & 75.00 $\pm$ 0.00 & \textbf{100.00 $\pm$ 0.00} & 5.00 \\
 \emph{RepNodes (best)} & - & \textbf{75.45 $\pm$ 1.90} & 61.43 $\pm$ 3.91 & - & \textbf{91.67 $\pm$ 0.00} & - & \textbf{1.50} \\
 \emph{RepEdges (SRL*)} & \underline{70.59 $\pm$ 0.00} & 71.82 $\pm$ 3.45 & 61.43 $\pm$ 3.91 & \textbf{100.00 $\pm$ 0.00} & 83.33 $\pm$ 0.00 & \textbf{100.00 $\pm$ 0.00} & 3.50 \\
 \emph{RepEdges (best)} & - & \textbf{75.45 $\pm$ 1.90} & - & - & \textbf{91.67 $\pm$ 0.00} & - & \underline{1.83} \\
\hline
\end{tabular}}
\caption{GCN node classification test accuracies on \emph{ECHO} datasets. For RAwR, we report both the result for the epsilon chosen by a grid search on validation set and the one chosen by the SRL* heuristic; if they coincide we report '-'. OOM stands for "Out Of Memory". Best result per dataset is \textbf{bold}, second-best is \underline{underlined}. We also report the Average Rank (lower is better).
}
\label{tab:cyhmn_gcn_srlstar}
\end{table}

\begin{table*}[b]
\centering
\scalebox{0.7}{
\begin{tabular}{lccccccccc}
\toprule
 & Actor & Chameleon & Citeseer & Cora & Cornell & PubMed & Squirrel & Texas & Wisconsin \\
\midrule
GCN & -0.18 & \cellcolor{green!20}\textbf{+11.54} & \cellcolor{green!20}\textbf{+3.61} & \cellcolor{green!20}\textbf{+1.00} & \textbf{-2.22} & \cellcolor{green!20}+0.51 & \cellcolor{green!20}\textbf{+7.08} & \cellcolor{green!20}\textbf{+10.00} & \cellcolor{green!20}\textbf{+1.60} \\
\bottomrule
\end{tabular}
}
\caption{Difference in mean test accuracy between RAwR RepNodes and a control variant where the partition is replaced by a random assignment. We highlight in green positive differences and we report in bold values greater than $1$ in absolute value.}
\label{tab:approx_vs_random_best_delta_sameeps}
\end{table*}

\paragraph{RAwR reduces the mean effective resistance.}
First, in Figure \ref{fig:effRes} 
we show that RAwR reduces the mean effective resistance of the input graph. This reduction is visible on all real-world datasets and is even more pronounced on \emph{ECHO}. These results provide structural evidence that RAwR alleviates long-range propagation bottlenecks, independently of the downstream GNN architecture.
We highlight that, for the same $\varepsilon$-AEP we would expect RepEdges to always have less resistance than RepNodes: this is because RepEdges only differs by an addition of edges from the quotient graph that trivially decrease the mean effective resistance.
But in
we are showing the results for values of $\varepsilon$ that achieve the best node classification accuracy for each variant. Therefore, the two variants may correspond to different $\varepsilon$-AEP.
This explains why in Cornell, PubMed, Wisconsin and Lobster the mean effective resistance for RepEdges is higher than for RepNodes.

\paragraph{RAwR achieves the best average rank.}
The first main observation is that RAwR obtains the best average rank among all compared rewiring methods. On the real-world benchmarks (Table~\ref{tab:ncq1}), RepNodes with the best value of $\varepsilon$ achieves an average rank of $3.00$ both with and without node features, outperforming the vanilla baseline and all competing rewiring baselines. The same conclusion becomes even clearer on \emph{ECHO} (Table~\ref{tab:cyhmn_gcn_srlstar}). This indicates that the improvements are not tied to a specific dataset family, but persist across homophilic, heterophilic, and explicitly long-range settings. Moreover, while some rewiring methods obtain strong results on individual datasets, their rankings are less stable across regimes. RAwR is instead consistently among the strongest methods, suggesting that role-aware rewiring provides a robust criterion for modifying the graph topology.

\paragraph{RAwR works best for long-range and heterophilic benchmarks.}
On homophilic datasets such as Cora, Citeseer, and PubMed, the gains are generally moderate, as local aggregation is already well aligned with the labeling function. In these cases, RAwR usually remains close to the baseline and does not substantially disrupt local message passing. On heterophilic datasets, instead, the improvements are more pronounced, showing that role-aware shortcuts can help when immediate neighbors are not necessarily label-informative. The strongest evidence comes from \emph{ECHO}, where the task explicitly requires long-range propagation: here RAwR (in all feature regimes and for both RepNodes and RepEdges) achieves an Average Rank between 1.83 and 1.33, with substantial improvements on node classification accuracies (like in the Caterpillar dataset, where accuracies goes from $38.82$ to $76.47$ in the no-feature regime).

\paragraph{RepNodes vs. RepEdges.}
Comparing the two variants, RepNodes generally performs better than RepEdges, especially on the real-world benchmarks. This suggests that the main source of improvement comes from fast intra-role communication, obtained by connecting each node to its role representative. RepEdges remains competitive, and is particularly strong on the \emph{ECHO} benchmark.

\paragraph{On the selection of $\varepsilon$ via SRL$^*$.}
The SRL$^*$ heuristic provides a practical way to select the tolerance parameter $\varepsilon$. On the real-world benchmarks, the configurations selected by SRL$^*$ are often close to the best configurations found by grid search on the validation set, and RepNodes selected through SRL$^*$ still obtains the best average rank among the non-oracle RAwR variants. On \emph{ECHO}, the gap between SRL$^*$ and the best $\varepsilon$ is larger, especially for RepNodes, suggesting that synthetic long-range structures can stress the heuristic more severely. Nevertheless, SRL$^*$ remains informative enough to select competitive rewired graphs, while the best-$\varepsilon$ results show the full potential of the RAwR construction.

\paragraph{Comparison with random partitions.}
Table~\ref{tab:approx_vs_random_best_delta_sameeps} reports the absolute difference in mean test accuracy, measured in percentage points, between RAwR RepNodes and a matched random-partition control. The control uses the same number of virtual nodes and the same $\varepsilon$ regime as RAwR, but replaces the $\varepsilon$-AEP assignment with random node-to-block assignments. Therefore, positive values indicate the gain obtained by using role-aware partitions instead of arbitrary virtual-node connections.
For GCN, RAwR improves over the matched random control by an average of $+3.66$ percentage points across datasets. The largest gains appear on heterophilous graphs, e.g., Chameleon $(+11.54)$, Squirrel $(+7.08)$, and Texas $(+10.00)$, where local adjacency is less aligned with label similarity and the role assignment becomes more important. On more homophilous graphs, the differences are smaller, consistently with the fact that local message passing is already informative. Overall, this ablation shows that the gains of RAwR are not explained by merely adding the same number of virtual nodes, but by assigning them according to structural roles.

\section{Conclusions and Future Work}
\label{sec:conclusion}

We introduced RAwR, a computationally efficient and model-agnostic rewiring framework that augments the input graph with a quotient graph derived from approximate equitable partitions. The central idea is that rewiring should not only reduce propagation bottlenecks, but also route information through structurally meaningful paths. By assigning virtual representatives to structural roles, RAwR enables fast communication between nodes that are topologically distant but structurally similar, while recovering the Master Node construction as the fully collapsed limit of the partition.

Our experiments show that this role-aware routing is effective across homophilic, heterophilic, and synthetic long-range benchmarks. RAwR achieves the best average rank among the compared rewiring methods, with the strongest gains on heterophilic and long-range datasets, where local message passing is insufficient. The reduction in mean effective resistance further provides structural evidence that RAwR alleviates long-range propagation bottlenecks independently of the downstream GNN architecture. The random-partition control confirms that the improvements are driven by the approximate equitable partition, rather than by the mere addition of virtual nodes.

On the theoretical side, we supported the method through a teacher-student analysis of linear GNNs, showing how role-aware rewiring modifies the spectral response available to message-passing models. This led to Spectral Role Lift (SRL), and to the practical SRL$^*$ heuristic for selecting the tolerance parameter $\varepsilon$. Future work will extend RAwR to directed and weighted graphs, where approximate equitable partitions are already defined, and will further refine SRL-based selection to better account for the approximation error induced by coarser partitions.

\paragraph{Limitations.}
RAwR is currently evaluated on undirected and unweighted graphs. Although approximate equitable partitions are defined for directed and weighted graphs, the present SRL derivation assumes undirected normalized shifts. Moreover, the SRL heuristic is derived under a linear teacher-student model and may be less informative for architectures whose propagation cannot be easily represented as a polynomial filter. Finally, the empirical evaluation focuses on node classification; extending the analysis to other graph-learning tasks is left for future work.

\bibliographystyle{plainnat}
\bibliography{neurips_2026}


\appendix


\newpage

\section{EP and AEP extended background}
\label{app:extended_background}

In this appendix we show the exact and approximate equitable partition computed on a simple running example. 

\begin{figure}[h]
\centering
\begin{subfigure}{.30\columnwidth}
	\centering
 \includegraphics[width=\textwidth]{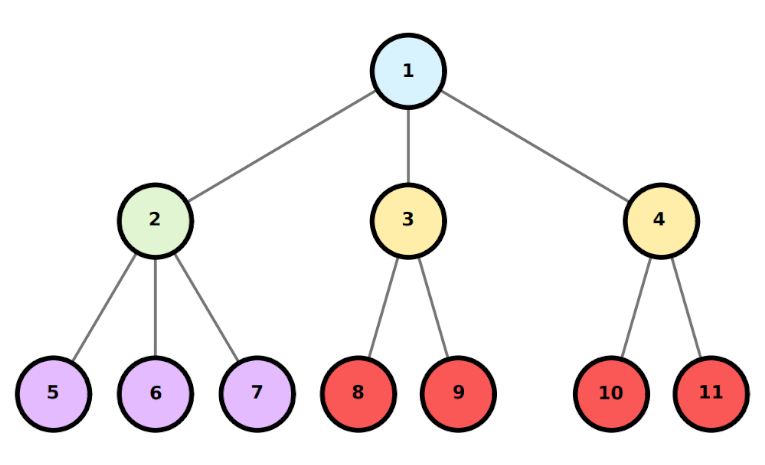}
	\caption{}
	\label{fig:exact}
\end{subfigure}
\hspace{1.5cm}
\begin{subfigure}{.30\columnwidth}
        \centering
 \includegraphics[width=\textwidth]{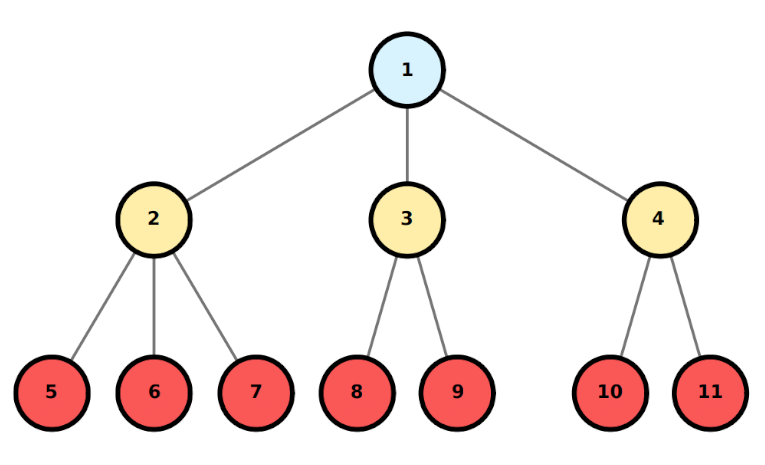}
	\caption{}
	\label{fig:approx}
\end{subfigure}
\caption{(a) A graph with a color-coding denoting its exact EP. (b) The AEP obtained for $\varepsilon=1$.}
\label{fig:netRed}
\end{figure}

Figure~\ref{fig:netRed} (a) shows a graph and its equitable partition. Here nodes with the same colors correspond to the same structural role in the network. This role is characterized by a specific number of connection for each block. For example, node $3$ and $4$ are connected with 2 red nodes an 1 cyan node.

Figure~\ref{fig:netRed} (b), shows a graph and its approximated equitable partition obtained for $\varepsilon=1$. The color coding shows that the 5 different structural roles got collapsed into 3. In this case, nodes assigned to the same role exhibit similar, though not identical, connection patterns with differences bounded by the tolerance $\varepsilon$, resulting in a less fine-grained partition of the network. 

\section{Quantifying the impact of RAwR. Detailed derivation of SRL}
\label{app:roles_important}

This appendix provides the formal derivation of Eq.~\eqref{eq:srl_err_lin_relation}
starting from the teacher--student setting introduced in Section~\ref{sec:ts_model}.
We follow the role-wise reduction and energy decomposition developed in the
supplementary derivation.

\subsection{From Section~\ref{sec:ts_model} to a shared effective signal}
In Section~\ref{sec:ts_model} we defined
\(
y_{\text{true}} = \big(h(S_{\mathrm{RAwR}}) X_{\mathrm{RAwR}} \prod_{l=1}^L W_t^{(l)}\big)_{1:n,:}
\)
and
\(
y_{\text{obs}}(W_s) = h(S_{\mathrm{obs}}) X \prod_{l=1}^L W_s^{(l)}.
\)
Since linear GNNs trained with gradient descent converge to a global optimum,
we have \(\mathrm{MSE}(W_s^\star)\le \mathrm{MSE}(W_t)\) and it suffices to upper bound
\(\mathrm{MSE}(W_t)\) (cf.\ Section~\ref{sec:ts_model}).  

Define the effective (teacher) signal
\begin{equation*}
B := \big(X_{\mathrm{RAwR}} \prod_{l=1}^L W_t^{(l)}\big)_{1:n,:}
\end{equation*}
and denote by \(b_c\in\mathbb{R}^n\) its \(c\)-th column.
In the derivation below we compare teacher and student outputs obtained by applying the
\emph{same} effective signal \(B\) to two different shifts \(S_{\mathrm{RAwR}}\) and \(S_{\mathrm{obs}}\),
which is exactly the comparison implicit in \(\mathrm{MSE}(W_t)\). 
In the following section we will define useful quantities in sections \ref{app:def_shifts}, \ref{app:def_per_role} and \ref{app:def_role_eigenbasis} which will be used in the calculation of the upper bound on the error in section \ref{app:error_bound_calc_start}.

\subsection{Shifts, equitable partition, and role basis}
\label{app:def_shifts}
Let \(A\) be the observed adjacency and \(S_{\mathrm{obs}}=D_{\mathrm{obs}}^{-1/2}(A+I)D_{\mathrm{obs}}^{-1/2}\)
the normalized shift with self-loops. Let \(\mathcal{B}=\{B_1,\dots,B_k\}\) be an equitable partition,
\(R\in\mathbb{R}^{n\times k}\) the node--block incidence matrix, and
\begin{equation*}
C := R(R^\top R)^{-1/2}\in\mathbb{R}^{n\times k},\qquad C^\top C = I_k,
\qquad P_U := CC^\top,
\end{equation*}
so that \(U:=\mathrm{im}(C)\) is the role subspace and \(P_U\) its orthogonal projector.

Under exact EP, the augmented adjacency is
\begin{equation*}
A_{\mathrm{RAwR}} =
\begin{pmatrix}
A & R\\
R^\top & Q
\end{pmatrix},
\end{equation*}
with quotient matrix \(Q\), and the corresponding normalized shift on the augmented graph satisfies
\begin{equation*}
S_{\mathrm{RAwR}}
=
D_{\mathrm{RAwR}}^{-1/2}(A_{\mathrm{RAwR}}+I)D_{\mathrm{RAwR}}^{-1/2}
=
\begin{pmatrix}
S_{oo} & S_{ov}\\
S_{vo} & S_{vv}
\end{pmatrix}.
\end{equation*}
In general \(S_{oo}\neq S_{\mathrm{obs}}\) because degrees of observed nodes change after augmentation.

\subsection{Per-role reduction and the lifted eigenvalues \(\lambda_{+,j}\)}
\label{app:def_per_role}
For each role direction \(C_j\) (the \(j\)-th column of \(C\)), define
\begin{equation*}
\mu^{\mathrm{obs}}_j := \langle C_j, S_{\mathrm{obs}} C_j\rangle,
\qquad
\mu^{\mathrm{RAwR}}_j := \langle C_j, S_{oo} C_j\rangle,
\qquad
\tau_j := \|S_{vo}C_j\|.
\end{equation*}
Set \(\widehat v_j := S_{vo}C_j/\tau_j\) and \(\nu_j := \langle \widehat v_j, S_{vv}\widehat v_j\rangle\).
Consider the \(2\times 2\) symmetric matrix
\begin{equation*}
M_j :=
\begin{pmatrix}
\mu^{\mathrm{RAwR}}_j & \tau_j\\
\tau_j & \nu_j
\end{pmatrix},
\end{equation*}
and denote by \(\lambda_{\pm,j}\) its eigenvalues.
We focus on the positive branch \(\lambda_{+,j}\) (since GNNs are known to act as low-pass filters) and define the spectral lift
\begin{equation*}
\Delta_j := \lambda_{+,j}-\mu^{\mathrm{obs}}_j \ge 0.
\end{equation*}
This \(\Delta_j\) is exactly the ``difference of role-relevant eigenvalues'' referenced in Section~\ref{sec:ts_model}. 

\subsection{Role--eigenbasis and scalar responses of the filter}
\label{app:def_role_eigenbasis}
Under the EP assumption, restricting the shifts to \(U\) yields \(k\times k\) symmetric matrices.
Hence, if the restricted shifts commute, there exists an orthogonal \(V\) such that both restrictions are jointly diagonalized;
without loss of generality we assume \(C\leftarrow CV\) and still denote it by \(C\).
In this role--eigenbasis we have (cf.\ Section~\ref{sec:ts_model})
\begin{equation*}
C^\top S_{\mathrm{RAwR}} C = \Lambda^{(U)}_{\mathrm{RAwR}}=\mathrm{diag}(\lambda_{+,1},\dots,\lambda_{+,k}),
\qquad
C^\top S_{\mathrm{obs}} C = \Lambda^{(U)}_{\mathrm{obs}}=\mathrm{diag}(\mu^{\mathrm{obs}}_1,\dots,\mu^{\mathrm{obs}}_k),
\end{equation*}
and since \(h\) is a polynomial the same diagonalization holds for \(h(S)\) restricted to \(U\):
\begin{equation*}
C^\top h(S_{\mathrm{RAwR}}) C = h(\Lambda^{(U)}_{\mathrm{RAwR}}),
\qquad
C^\top h(S_{\mathrm{obs}}) C = h(\Lambda^{(U)}_{\mathrm{obs}}).
\end{equation*}
In particular, for each role direction \(C_j\) and every \(v\),
\begin{equation*}
C_j^\top h(S_{\mathrm{RAwR}})v = h(\lambda_{+,j})\, C_j^\top v,
\qquad
C_j^\top h(S_{\mathrm{obs}})v = h(\mu^{\mathrm{obs}}_j)\, C_j^\top v.
\end{equation*} 

While the commutativity of the restricted shifts operators cannot be guaranteed on principle, there are works proving that almost commutative matrices are also almost jointly diagonalizable \cite{DBLP:journals/corr/abs-1305-2135}.
To support this assumption empirically, Table~\ref{tab:comm_norm} reports the Frobenius norm of the commutator between the restricted shifts,
computed for all datasets and all values of $\varepsilon$ used in the experiments (cf. Section \ref{sec:exps}).

\begin{table}[t]
\centering
\begin{tabular}{lccccc}
\toprule
Dataset &
$0^\circ$ (Exact EP)&
$25^\circ$ &
$50^\circ$ &
$75^\circ$ &
$100^\circ$ (MN)\\
\midrule
Actor      & $1.0\times10^{-3}$ & $1.0\times10^{-2}$ & $3.3\times10^{-2}$ & $5.3\times10^{-2}$ & $0$ \\
Citeseer   & $1.6\times10^{-3}$ & $3.6\times10^{-3}$ & $9.4\times10^{-3}$ & $2.4\times10^{-2}$ & $0$ \\
Cora       & $1.4\times10^{-3}$ & $1.4\times10^{-2}$ & $4.0\times10^{-2}$ & $5.8\times10^{-2}$ & $0$ \\
Texas      & $6.8\times10^{-3}$ & $3.1\times10^{-2}$ & $5.5\times10^{-2}$ & $6.6\times10^{-2}$ & $0$ \\
Wisconsin  & $5.5\times10^{-3}$ & $1.6\times10^{-2}$ & $3.1\times10^{-2}$ & $6.5\times10^{-2}$ & $0$ \\
\bottomrule
\end{tabular}
\caption{Frobenius norm of the commutator of the restriced shifts operators for values of $\varepsilon$ selected from the percentiles of the degree sequence of the observed graph.
The 0th-percentile corresponds to the exact EP and the 100th-percentile corresponds to the Master Node.}
\label{tab:comm_norm}
\end{table}

\subsection{Role-wise error for a single output dimension}
\label{app:error_bound_calc_start}
Let \(y^{(c)}_{\text{true}}\) and \(y^{(c)}_{\text{obs}}\) denote the \(c\)-th output columns (restricted to real nodes),
and define the class-wise error consistently with Section~\ref{sec:ts_model} as
\begin{equation*}
e^{(c)} := y^{(c)}_{\text{true}} - y^{(c)}_{\text{obs}}
= \big(h(S_{\mathrm{RAwR}})-h(S_{\mathrm{obs}})\big)b_c.
\end{equation*}
Define the role coefficients (teacher and observed)
\begin{equation*}
\beta^{\mathrm{true}}_{j,c} := C_j^\top y^{(c)}_{\text{true}},\qquad
\beta^{\mathrm{obs}}_{j,c} := C_j^\top y^{(c)}_{\text{obs}},\qquad
\alpha_{j,c}:=C_j^\top b_c.
\end{equation*}
Applying the scalar responses above with \(v=b_c\) gives
\begin{equation} \label{eq:beta_scalar_alpha}
\beta^{\mathrm{true}}_{j,c} 
\simeq h(\lambda_{+,j})\,\alpha_{j,c},\qquad
\beta^{\mathrm{obs}}_{j,c} \simeq h(\mu^{\mathrm{obs}}_j)\,\alpha_{j,c},
\end{equation}
hence the role-wise error coefficient is
\begin{equation*}
\varepsilon_{j,c}:=C_j^\top e^{(c)}=\beta^{\mathrm{true}}_{j,c}-\beta^{\mathrm{obs}}_{j,c}
\simeq \big(h(\lambda_{+,j})-h(\mu^{\mathrm{obs}}_j)\big)\alpha_{j,c}.
\end{equation*}
The $\simeq$ symbol is due to the fact that, as stated in \ref{app:def_role_eigenbasis}, exact commutativity of the restricted shift operators is not guaranteed.

Since \(\|C_j\|=1\), the squared error contribution of role \(j\) for class \(c\) is
\begin{equation*}
\|e^{(c)}_j\|^2 := \|\varepsilon_{j,c} C_j\|^2 = \varepsilon_{j,c}^2
 \simeq \big(h(\lambda_{+,j})-h(\mu^{\mathrm{obs}}_j)\big)^2\alpha_{j,c}^2.
\end{equation*}

Using Eq.~\eqref{eq:beta_scalar_alpha}, and separating the case \(h(\lambda_{+,j})=0\),
we obtain exactly the decomposition used throughout the paper:
\begin{equation}
\label{eq:role_error_decomp_appendix}
\|e^{(c)}_j\|^2 =
\begin{cases}
(\beta^{\mathrm{obs}}_{j,c})^2 & \text{if } h(\lambda_{+,j})=0,\\[3pt]
\dfrac{\big(h(\mu^{\mathrm{obs}}_j)-h(\lambda_{+,j})\big)^2}{h(\lambda_{+,j})^2}\,(\beta^{\mathrm{true}}_{j,c})^2
& \text{if } h(\lambda_{+,j})\neq 0.
\end{cases}
\end{equation}

\subsection{Mean value theorem and the \(\kappa\)-constants}
For \(h(\lambda_{+,j})\neq 0\), since \(h\) is differentiable on the spectral interval,
the mean value theorem guarantees that for each \(j\) there exists \(\xi_j\) between
\(\mu^{\mathrm{obs}}_j\) and \(\lambda_{+,j}\) such that
\begin{equation*}
h(\lambda_{+,j})-h(\mu^{\mathrm{obs}}_j)=h'(\xi_j)\,(\lambda_{+,j}-\mu^{\mathrm{obs}}_j)
=h'(\xi_j)\,\Delta_j.
\end{equation*}
Therefore,
\begin{equation} \label{eq:mvt_upper}
\|e^{(c)}_j\|^2 \simeq \kappa_j\,\Delta_j^2\,(\beta^{\mathrm{true}}_{j,c})^2,
\qquad
\kappa_j := \frac{h'(\xi_j)^2}{h(\lambda_{+,j})^2},
\qquad
\kappa_{\max}:=\max_{j:\,h(\lambda_{+,j})\neq 0}\kappa_j.
\end{equation} 

\subsection{Energy fractions and class-wise SRL}
For each output dimension \(c\), define the total energy
\begin{equation*}
E_c := \|y^{(c)}_{\text{true}}\|^2,
\end{equation*}
and the portion lying in the role subspace
\begin{equation*}
E^{\mathrm{roles}}_c := \|P_U y^{(c)}_{\text{true}}\|^2 = \sum_{j=1}^k (\beta^{\mathrm{true}}_{j,c})^2.
\end{equation*}
Define the fraction of energy in the role subspace
\begin{equation*}
\rho_c := \frac{E^{\mathrm{roles}}_c}{E_c},
\end{equation*}
and the role-importance weights
\begin{equation*}
\omega_{j,c} := \frac{(\beta^{\mathrm{true}}_{j,c})^2}{\sum_{h=1}^k(\beta^{\mathrm{true}}_{h,c})^2},
\qquad \sum_{j=1}^k \omega_{j,c}=1.
\end{equation*}
These definitions are purely algebraic and imply
\begin{equation*}
(\beta^{\mathrm{true}}_{j,c})^2 = \omega_{j,c}\,E^{\mathrm{roles}}_c
= \omega_{j,c}\,\rho_c\,E_c.
\end{equation*} 

Substituting into Eq.~\eqref{eq:role_error_decomp_appendix}, applying Eq.~\eqref{eq:mvt_upper} and summing over roles yields the class-wise upper bound
\begin{equation}
\label{eq:classwise_bound_appendix}
\sum_{j=1}^k \|e^{(c)}_j\|^2
\lesssim
\sum_{j:\,h(\lambda_{+,j})=0} (\beta^{\mathrm{obs}}_{j,c})^2
\;+\;
\kappa_{\max}\,\rho_c\,E_c \sum_{j:\,h(\lambda_{+,j})\neq 0}\omega_{j,c}\,\Delta_j^2.
\end{equation} 

Motivated by the second term, we define the class-wise spectral role lift
\begin{equation*}
\mathrm{SRL}_c := \rho_c \sum_{j=1}^k \omega_{j,c}\,\Delta_j^2,
\end{equation*}
which is the quantity that appears in the role-wise MSE bound.

\subsection{From class-wise to global SRL}
Define the total label energy and its role component as
\begin{equation*}
E_{\mathrm{tot}} := \sum_c E_c,
\qquad
E^{\mathrm{roles}}_{\mathrm{tot}} := \sum_c E^{\mathrm{roles}}_c
= \sum_c\sum_{j=1}^k (\beta^{\mathrm{true}}_{j,c})^2,
\end{equation*}
and the global fraction of energy in the role subspace
\begin{equation*}
\rho := \frac{E^{\mathrm{roles}}_{\mathrm{tot}}}{E_{\mathrm{tot}}}.
\end{equation*}
Define also the global role weights
\begin{equation*}
\omega_j := \frac{\sum_c(\beta^{\mathrm{true}}_{j,c})^2}{\sum_{h=1}^k\sum_c(\beta^{\mathrm{true}}_{h,c})^2},
\qquad \sum_{j=1}^k \omega_j = 1.
\end{equation*}
Then the global SRL is
\begin{equation*}
\label{eq:srl_global_quadratic_appendix}
\mathrm{SRL} := \rho \sum_{j=1}^k \omega_j\,\Delta_j^2,
\end{equation*}
which coincides with the energy-weighted average of \(\mathrm{SRL}_c\) across classes. 

Finally, summing Eq.~\eqref{eq:classwise_bound_appendix} over \(c\) and defining
\begin{equation*}
\kappa_0 := \sum_c \sum_{j:\,h(\lambda_{+,j})=0}(\beta^{\mathrm{obs}}_{j,c})^2,
\end{equation*}
we obtain
\begin{equation*}
\sum_c \|e^{(c)}\|^2
\lesssim
\kappa_0 + \kappa_{\max} E_{\mathrm{tot}}\,\mathrm{SRL},
\end{equation*}
which is exactly Eq.~\eqref{eq:srl_err_lin_relation} in Section~\ref{sec:ts_model}.

\section{Details on teacher--student experiment}
\label{app:details_ts_experiment}
For the results in Figure~\ref{fig:ts_experiment} we consider a 2-layer standard GCN.
The graph structure and input features are taken from the real-world datasets.
To generate the teacher outputs we drew one sample for the weights from a Gaussian i.i.d prior, i.e., $W^{(l)}_{ij} \sim \mathcal{N}(0, \sigma_l^2/d_{prev})$ where $d_{prev}$ denotes the feature dimension of the previous layer.
For Figure~\ref{fig:ts_experiment} we chose $\sigma_2=40$ as well as $\sigma_1=\sigma_0=1$.
The output dimension has been chosen to match the number of classes of the respective real-world dataset.
Training has been done with the Adam optimizer with learning rate of $0.005$ for $5000$ epochs using full batches.

\section{SRL* and test accuracies}
\label{app:srl_corr}

Overall, $\mathrm{SRL}^*$ exhibits a strong correlation with downstream performance in several settings, particularly for GCNs on heterophilous datasets such as Chameleon, Squirrel, and Wisconsin, supporting its use as a performance proxy.
On homophilous datasets (e.g., Cora and Citeseer), the correlation is weaker or negative, reflecting the limited impact of rewiring when local message passing is already well aligned with the labeling function and, in some cases, the presence of outliers that induce negative correlations (see Appendix \ref{app:srl_corr}).

\begin{table*}[h]
\centering
\label{tab:srl*corr}
\resizebox{\textwidth}{!}{%
\begin{tabular}{lllccccccccc}
\toprule
 &  & Dataset & Actor & Chameleon & Citeseer & Cora & Cornell & PubMed & Squirrel & Texas & Wisconsin \\
Model & Features & Method &  &  &  &  &  &  &  &  &  \\
\midrule
\multirow[t]{4}{*}{GCN} & \multirow[t]{2}{*}{Yes} & RepNodes & \textbf{0.79} & \textbf{0.98} & -0.74 & -0.69 & \textbf{0.91} & 0.06 & \textbf{0.99} & -0.24 & \textbf{0.99} \\
 &  & RepEdges & 0.33 & \textbf{0.94} & -0.75 & -0.76 & \textbf{0.69} & 0.06 & \textbf{0.98} & -0.32 & \textbf{0.97} \\
\cmidrule(lr){2-12}
 & \multirow[t]{2}{*}{No} & RepNodes & 0.13 & \textbf{1.00} & \textbf{0.88} & \textbf{0.61} & \textbf{0.86} & \textbf{0.67} & \textbf{1.00} & -0.29 & \textbf{0.93} \\
 &  & RepEdges & -0.05 & \textbf{0.92} & -0.80 & \textbf{0.66} & \textbf{0.77} & 0.16 & \textbf{0.96} & -0.69 & -0.25 \\
\midrule
\multirow[t]{4}{*}{GAT} & \multirow[t]{2}{*}{Yes} & RepNodes & 0.59 & \textbf{0.72} & -0.55 & 0.09 & 0.37 & -0.41 & -0.97 & -0.46 & 0.35 \\
 &  & RepEdges & 0.59 & \textbf{0.64} & -0.41 & 0.06 & 0.36 & -0.41 & -0.95 & -0.41 & 0.35 \\
\cmidrule(lr){2-12}
 & \multirow[t]{2}{*}{No} & RepNodes & \textbf{0.82} & \textbf{0.66} & -0.42 & -0.94 & -- & -0.36 & -0.53 & -0.59 & 0.52 \\
 &  & RepEdges & \textbf{0.83} & 0.58 & -0.33 & -0.93 & -- & -0.37 & -0.61 & -0.61 & 0.50 \\
\midrule
\multirow[t]{4}{*}{GIN} & \multirow[t]{2}{*}{Yes} & RepNodes & \textbf{0.89} & \textbf{0.97} & -0.02 & 0.51 & 0.46 & -0.81 & 0.49 & -0.97 & -0.26 \\
 &  & RepEdges & \textbf{0.89} & \textbf{0.93} & -0.06 & 0.47 & 0.48 & -0.80 & -0.99 & -0.95 & -0.29 \\
\cmidrule(lr){2-12}
 & \multirow[t]{2}{*}{No} & RepNodes & 0.14 & \textbf{0.92} & 0.05 & 0.22 & -0.11 & 0.28 & \textbf{0.74} & -0.97 & 0.24 \\
 &  & RepEdges & 0.13 & \textbf{0.93} & 0.06 & 0.19 & -0.09 & 0.48 & -1.00 & -0.98 & 0.20 \\
\bottomrule
\end{tabular}%
}
\caption{Pearson correlation (across $\varepsilon$) between test accuracies and SRL* for node classification at $L=2$. Correlations greater than $0.6$ are in bold.}
\end{table*}

\section{Further details on experimental setup and additional experiments}
\label{app:additional_experiments}
We give here additional details on the experimental setup which were not given in Section \ref{sec:exps}.

For the three backbone models, we try a set $\{2,3,4,5,6,7,8\}$ of model depths (number of message passing layers).
Adam optimizer is adopted with learning rate set to $0.01$, $500$ epochs and early stopping based on the validation metric. Each configuration (dataset, GNN model, feature presence, augmentation variant, $\varepsilon$, model depth) is run five times with different random splits.
We optimize cross-entropy loss on labeled nodes and report the node classification test accuracy.
The train/val/test split ratios are, respectively, $85\%$, $5\%$ and $10\%$.

Below, we show the results for $8 \geq L > 2$. In Table \ref{tab:ncq3}, we take the model depth for which the baseline obtains the best accuracy and we train the backbone models on graphs with MN and RAwR variants by fixing the same number of GNN layers.
In Table \ref{tab:ncq4}, each row is evaluated at the model depth for which the GNN obtained the best node classification accuracy.
RAwR consinstently outperforms MN and baselines across all depths.

\begin{table}[H]
\centering
\resizebox{\textwidth}{!}{%
\begin{tabular}{llccccccccc c}
\toprule
 &  & Actor & Chameleon & Citeseer & Cora & Cornell & PubMed & Squirrel & Texas & Wisconsin & Avg.Rank \\
\cmidrule{1-11} \cmidrule(l){12-12}
\multicolumn{12}{c}{\emph{WITH NODE FEATURES}}
\\
\cmidrule{1-11} \cmidrule(l){12-12}
\multicolumn{12}{l}{\textbf{GCN}} \\
 & Baseline & 25.00 $\pm$ 0.74 & 65.02 $\pm$ 0.59 & 75.18 $\pm$ 0.59 & 82.00 $\pm$ 0.97 & 44.44 $\pm$ 0.00 & 86.15 $\pm$ 0.04 & 61.65 $\pm$ 0.22 & 33.33 $\pm$ 0.00 & 52.00 $\pm$ 0.00 & 7.44 \\
 & BORF & 25.63 $\pm$ 0.23 & 66.43 $\pm$ 0.70 & 68.92 $\pm$ 0.35 & 82.96 $\pm$ 0.00 & 45.56 $\pm$ 4.16 & OOM & OOM & 36.67 $\pm$ 6.67 & 56.80 $\pm$ 2.99 & 7.22 \\
 & FOSR & 25.50 $\pm$ 0.38 & 65.99 $\pm$ 0.33 & 68.49 $\pm$ 0.24 & 83.11 $\pm$ 0.18 & 44.44 $\pm$ 0.00 & 82.09 $\pm$ 0.06 & 61.77 $\pm$ 0.54 & 33.33 $\pm$ 0.00 & 55.20 $\pm$ 5.88 & 7.22 \\
 & SDRF & 25.37 $\pm$ 0.44 & 64.85 $\pm$ 0.58 & 69.04 $\pm$ 0.23 & 83.48 $\pm$ 0.18 & 44.44 $\pm$ 0.00 & OOM & 61.65 $\pm$ 0.31 & 33.33 $\pm$ 0.00 & 51.20 $\pm$ 1.60 & 8.00 \\
 & JDR & 29.16 $\pm$ 0.11 & 66.52 $\pm$ 1.53 & 74.52 $\pm$ 0.53 & \underline{83.70 $\pm$ 0.41} & 41.11 $\pm$ 2.72 & \underline{87.02 $\pm$ 0.09} & 53.19 $\pm$ 0.67 & \underline{46.67 $\pm$ 2.72} & \underline{69.60 $\pm$ 1.96} & 4.78 \\
 & ComFy & \underline{29.37 $\pm$ 0.75} & 64.32 $\pm$ 0.54 & \textbf{78.19 $\pm$ 0.50} & \textbf{87.26 $\pm$ 0.77} & 31.11 $\pm$ 3.04 & \textbf{87.95 $\pm$ 0.08} & 42.04 $\pm$ 0.52 & 30.00 $\pm$ 4.97 & 53.60 $\pm$ 2.19 & 6.33 \\
 & TRIGON & \textbf{36.18 $\pm$ 1.87} & 29.43 $\pm$ 5.19 & 71.57 $\pm$ 2.05 & 78.00 $\pm$ 6.07 & \textbf{71.11 $\pm$ 2.48} & 80.73 $\pm$ 0.18 & 41.85 $\pm$ 1.16 & \textbf{70.00 $\pm$ 6.33} & \textbf{76.80 $\pm$ 3.35} & 6.22 \\
 & MN & 25.16 $\pm$ 0.68 & 66.30 $\pm$ 0.84 & OOM & 81.70 $\pm$ 0.40 & 47.78 $\pm$ 2.87 & OOM & 61.17 $\pm$ 0.29 & 34.44 $\pm$ 2.34 & 48.80 $\pm$ 1.69 & 9.00 \\[0.15cm]
 & \emph{RepNodes (SRL*)} & 25.95 $\pm$ 0.37 & \textbf{73.83 $\pm$ 0.50} & 74.58 $\pm$ 0.74 & 81.56 $\pm$ 0.41 & \underline{56.67 $\pm$ 2.48} & 86.34 $\pm$ 0.20 & \textbf{65.77 $\pm$ 0.38} & 33.33 $\pm$ 0.00 & 56.00 $\pm$ 4.00 & \underline{4.33} \\
 & \emph{RepNodes (best)} & - & - & \underline{75.24 $\pm$ 0.37} & 81.93 $\pm$ 0.66 & - & 86.46 $\pm$ 0.22 & - & 43.33 $\pm$ 2.48 & - & \textbf{3.00} \\
 & \emph{RepEdges (SRL*)} & 25.50 $\pm$ 0.24 & \underline{71.37 $\pm$ 0.31} & 74.58 $\pm$ 0.74 & 81.48 $\pm$ 0.45 & 50.00 $\pm$ 0.00 & 86.34 $\pm$ 0.20 & \underline{63.46 $\pm$ 0.19} & 33.33 $\pm$ 0.00 & 51.20 $\pm$ 3.35 & 6.11 \\
 & \emph{RepEdges (best)} & 25.68 $\pm$ 0.14 & - & \underline{75.24 $\pm$ 0.37} & 81.93 $\pm$ 0.66 & - & 86.46 $\pm$ 0.22 & - & 38.89 $\pm$ 5.56 & - & 4.56 \\
\cmidrule{1-11} \cmidrule(l){12-12}
\multicolumn{12}{l}{\textbf{GAT}} \\
 & Baseline & 25.39 $\pm$ 1.02 & \underline{60.09 $\pm$ 5.47} & 73.25 $\pm$ 0.97 & 81.85 $\pm$ 1.59 & 44.44 $\pm$ 3.93 & 85.43 $\pm$ 0.80 & 42.35 $\pm$ 2.82 & 41.11 $\pm$ 3.04 & 50.40 $\pm$ 6.69 & 5.56 \\
 & BORF & 26.11 $\pm$ 0.34 & 57.53 $\pm$ 2.45 & 60.18 $\pm$ 1.05 & 74.37 $\pm$ 2.60 & 44.44 $\pm$ 3.51 & OOM & OOM & \underline{46.67 $\pm$ 6.67} & 52.80 $\pm$ 6.88 & 7.89 \\
 & FOSR & 24.68 $\pm$ 1.28 & 59.65 $\pm$ 2.68 & 60.66 $\pm$ 1.85 & 74.89 $\pm$ 2.45 & 43.33 $\pm$ 2.22 & 77.13 $\pm$ 0.66 & \underline{45.00 $\pm$ 3.61} & 36.67 $\pm$ 6.67 & 50.40 $\pm$ 5.43 & 8.00 \\
 & SDRF & 24.53 $\pm$ 1.03 & 59.65 $\pm$ 3.84 & 60.06 $\pm$ 1.41 & 73.41 $\pm$ 1.00 & 42.22 $\pm$ 4.44 & OOM & 44.96 $\pm$ 3.12 & 36.67 $\pm$ 2.72 & 53.60 $\pm$ 8.62 & 8.78 \\
 & JDR & \underline{31.08 $\pm$ 0.87} & 50.66 $\pm$ 2.77 & 68.80 $\pm$ 0.82 & 71.70 $\pm$ 1.32 & 41.11 $\pm$ 6.67 & 75.79 $\pm$ 0.93 & 34.50 $\pm$ 2.30 & 42.22 $\pm$ 10.30 & \underline{64.00 $\pm$ 3.58} & 7.78 \\
 & ComFy & 28.89 $\pm$ 1.35 & 55.42 $\pm$ 7.04 & \textbf{77.23 $\pm$ 1.03} & \textbf{86.30 $\pm$ 1.05} & \underline{55.56 $\pm$ 5.56} & \textbf{85.79 $\pm$ 0.33} & 40.08 $\pm$ 3.22 & 34.44 $\pm$ 7.24 & 50.40 $\pm$ 2.19 & 5.44 \\
 & TRIGON & \textbf{36.34 $\pm$ 1.97} & 41.67 $\pm$ 8.03 & 68.92 $\pm$ 1.87 & 72.22 $\pm$ 2.81 & \textbf{71.11 $\pm$ 2.48} & 79.17 $\pm$ 0.84 & 40.85 $\pm$ 1.40 & \textbf{53.33 $\pm$ 10.09} & \textbf{79.20 $\pm$ 3.35} & 5.56 \\
 & MN & 23.39 $\pm$ 1.28 & 60.00 $\pm$ 5.20 & OOM & 80.74 $\pm$ 0.55 & 44.44 $\pm$ 0.00 & OOM & \textbf{45.77 $\pm$ 0.88} & 36.67 $\pm$ 2.87 & 48.80 $\pm$ 6.75 & 8.22 \\[0.15cm]
 & \emph{RepNodes (SRL*)} & 25.05 $\pm$ 0.80 & \textbf{66.43 $\pm$ 3.35} & 72.95 $\pm$ 1.05 & 82.15 $\pm$ 1.47 & 45.56 $\pm$ 2.48 & 85.42 $\pm$ 0.44 & 41.77 $\pm$ 1.07 & 35.56 $\pm$ 3.04 & 54.40 $\pm$ 7.80 & \underline{4.89} \\
 & \emph{RepNodes (best)} & - & - & \underline{73.61 $\pm$ 1.57} & \underline{82.44 $\pm$ 0.89} & - & \underline{85.46 $\pm$ 0.68} & \textbf{45.77 $\pm$ 0.93} & 41.11 $\pm$ 6.33 & - & \textbf{2.67} \\
 & \emph{RepEdges (SRL*)} & 25.05 $\pm$ 0.80 & \textbf{66.43 $\pm$ 3.35} & 72.95 $\pm$ 1.05 & 82.15 $\pm$ 1.47 & 45.56 $\pm$ 2.48 & 85.42 $\pm$ 0.44 & 41.77 $\pm$ 1.07 & 35.56 $\pm$ 3.04 & 54.40 $\pm$ 7.80 & \underline{4.89} \\
 & \emph{RepEdges (best)} & - & - & \underline{73.61 $\pm$ 1.57} & \underline{82.44 $\pm$ 0.89} & - & \underline{85.46 $\pm$ 0.68} & \textbf{45.77 $\pm$ 0.93} & 41.11 $\pm$ 6.33 & - & \textbf{2.67} \\
\cmidrule{1-11} \cmidrule(l){12-12}
\multicolumn{12}{l}{\textbf{GIN}} \\
 & Baseline & 24.39 $\pm$ 0.22 & 61.15 $\pm$ 2.56 & 73.13 $\pm$ 1.38 & 83.04 $\pm$ 1.03 & 51.11 $\pm$ 4.65 & 84.13 $\pm$ 0.35 & 65.23 $\pm$ 2.04 & \underline{58.89 $\pm$ 3.04} & 49.60 $\pm$ 6.07 & 6.67 \\
 & BORF & 24.84 $\pm$ 1.40 & 62.20 $\pm$ 1.75 & 67.83 $\pm$ 1.25 & 80.89 $\pm$ 1.94 & 48.89 $\pm$ 5.44 & OOM & OOM & 47.78 $\pm$ 2.72 & 57.60 $\pm$ 4.08 & 8.44 \\
 & FOSR & 24.34 $\pm$ 1.71 & 63.52 $\pm$ 1.40 & 67.11 $\pm$ 0.92 & 82.30 $\pm$ 0.86 & 44.44 $\pm$ 0.00 & 80.45 $\pm$ 0.40 & 63.38 $\pm$ 2.34 & 40.00 $\pm$ 4.16 & \underline{63.20 $\pm$ 2.99} & 7.78 \\
 & SDRF & 24.21 $\pm$ 1.93 & 63.44 $\pm$ 1.78 & 66.93 $\pm$ 0.44 & 82.22 $\pm$ 0.52 & 50.00 $\pm$ 6.09 & OOM & 64.00 $\pm$ 0.98 & 43.33 $\pm$ 8.16 & 55.20 $\pm$ 4.66 & 8.33 \\
 & JDR & 28.24 $\pm$ 1.76 & 40.97 $\pm$ 3.49 & 67.53 $\pm$ 1.05 & 73.04 $\pm$ 1.37 & 43.33 $\pm$ 4.16 & 79.74 $\pm$ 0.52 & 28.12 $\pm$ 3.91 & 56.67 $\pm$ 2.22 & 53.60 $\pm$ 4.80 & 8.11 \\
 & ComFy & \underline{28.37 $\pm$ 0.48} & \underline{63.96 $\pm$ 2.67} & \textbf{76.93 $\pm$ 1.39} & \textbf{86.00 $\pm$ 1.54} & 33.33 $\pm$ 6.80 & \textbf{86.05 $\pm$ 0.33} & 37.85 $\pm$ 2.91 & 34.44 $\pm$ 2.48 & 45.60 $\pm$ 5.37 & 6.22 \\
 & TRIGON & \textbf{38.58 $\pm$ 0.71} & 48.28 $\pm$ 5.65 & 69.58 $\pm$ 2.47 & 72.30 $\pm$ 8.33 & \textbf{57.78 $\pm$ 20.26} & 77.23 $\pm$ 0.26 & 42.04 $\pm$ 2.36 & 55.56 $\pm$ 13.03 & \textbf{74.40 $\pm$ 2.19} & 6.33 \\
 & MN & 24.92 $\pm$ 0.69 & 62.73 $\pm$ 3.09 & OOM & 82.74 $\pm$ 1.12 & 43.33 $\pm$ 2.34 & OOM & \underline{65.65 $\pm$ 3.58} & 56.67 $\pm$ 4.38 & 50.40 $\pm$ 3.37 & 7.78 \\[0.15cm]
 & \emph{RepNodes (SRL*)} & 25.66 $\pm$ 1.74 & \textbf{71.28 $\pm$ 1.57} & 74.52 $\pm$ 1.13 & \underline{83.93 $\pm$ 0.85} & 51.11 $\pm$ 7.24 & 84.50 $\pm$ 0.36 & \textbf{66.81 $\pm$ 3.30} & 35.56 $\pm$ 4.97 & 49.60 $\pm$ 4.56 & 4.33 \\
 & \emph{RepNodes (best)} & - & - & \underline{75.00 $\pm$ 0.70} & - & \underline{54.44 $\pm$ 7.24} & \underline{84.63 $\pm$ 0.39} & - & \textbf{61.11 $\pm$ 6.80} & 52.80 $\pm$ 1.79 & \textbf{2.33} \\
 & \emph{RepEdges (SRL*)} & 25.66 $\pm$ 1.74 & \textbf{71.28 $\pm$ 1.57} & 74.52 $\pm$ 1.13 & \underline{83.93 $\pm$ 0.85} & 51.11 $\pm$ 7.24 & 84.50 $\pm$ 0.36 & \underline{65.65 $\pm$ 3.80} & 35.56 $\pm$ 4.97 & 49.60 $\pm$ 4.56 & 4.56 \\
 & \emph{RepEdges (best)} & - & - & \underline{75.00 $\pm$ 0.70} & - & \underline{54.44 $\pm$ 7.24} & \underline{84.63 $\pm$ 0.39} & - & \textbf{61.11 $\pm$ 6.80} & 52.80 $\pm$ 1.79 & \underline{2.56} \\
\cmidrule{1-11} \cmidrule(l){12-12}
\multicolumn{12}{c}{\emph{WITHOUT NODE FEATURES}}
\\
\cmidrule{1-11} \cmidrule(l){12-12}
\multicolumn{12}{l}{\textbf{GCN}} \\
 & Baseline & 25.45 $\pm$ 0.78 & 67.84 $\pm$ 0.93 & 68.28 $\pm$ 0.20 & \underline{83.48 $\pm$ 0.19} & 44.44 $\pm$ 0.00 & 82.12 $\pm$ 0.06 & 60.85 $\pm$ 0.26 & 33.33 $\pm$ 0.00 & 50.40 $\pm$ 2.07 & 6.67 \\
 & BORF & 25.50 $\pm$ 0.92 & 68.46 $\pm$ 0.72 & 68.55 $\pm$ 0.24 & 83.19 $\pm$ 0.18 & 47.78 $\pm$ 2.72 & OOM & OOM & \underline{40.00 $\pm$ 6.48} & \textbf{62.40 $\pm$ 3.20} & 6.67 \\
 & FOSR & 25.82 $\pm$ 0.99 & 68.90 $\pm$ 0.72 & 68.31 $\pm$ 0.35 & 83.26 $\pm$ 0.15 & 44.44 $\pm$ 0.00 & 82.09 $\pm$ 0.06 & 61.19 $\pm$ 0.33 & 33.33 $\pm$ 0.00 & \underline{58.40 $\pm$ 1.96} & 5.78 \\
 & SDRF & 25.82 $\pm$ 0.97 & 67.93 $\pm$ 0.85 & 68.67 $\pm$ 0.33 & 83.33 $\pm$ 0.23 & 44.44 $\pm$ 0.00 & OOM & 61.31 $\pm$ 0.37 & 34.44 $\pm$ 2.22 & 52.00 $\pm$ 2.53 & 6.33 \\
 & JDR & \underline{25.95 $\pm$ 0.45} & 60.18 $\pm$ 0.77 & 69.46 $\pm$ 0.45 & 82.67 $\pm$ 0.28 & 44.44 $\pm$ 0.00 & \textbf{82.56 $\pm$ 0.12} & 51.08 $\pm$ 0.52 & 34.44 $\pm$ 2.22 & 50.40 $\pm$ 1.96 & 6.89 \\
 & ComFy & \textbf{28.32 $\pm$ 0.72} & 65.20 $\pm$ 0.82 & \textbf{72.53 $\pm$ 0.39} & \textbf{84.15 $\pm$ 0.17} & 28.89 $\pm$ 6.09 & \underline{82.15 $\pm$ 0.27} & 53.88 $\pm$ 0.76 & 27.78 $\pm$ 0.00 & 51.20 $\pm$ 4.38 & 6.00 \\
 & TRIGON & \textbf{28.32 $\pm$ 0.31} & 49.87 $\pm$ 12.91 & 56.39 $\pm$ 4.98 & 68.37 $\pm$ 7.29 & 6.67 $\pm$ 4.65 & 54.28 $\pm$ 3.33 & 54.31 $\pm$ 0.66 & 27.78 $\pm$ 0.00 & 54.40 $\pm$ 6.69 & 8.89 \\
 & MN & 25.39 $\pm$ 0.90 & 68.24 $\pm$ 0.35 & OOM & 83.24 $\pm$ 0.32 & 46.67 $\pm$ 2.79 & OOM & 60.73 $\pm$ 0.53 & 35.56 $\pm$ 2.79 & 49.60 $\pm$ 2.01 & 8.22 \\[0.15cm]
 & \emph{RepNodes (SRL*)} & 25.24 $\pm$ 0.68 & \textbf{75.07 $\pm$ 0.23} & \underline{70.84 $\pm$ 0.13} & 83.30 $\pm$ 0.27 & \textbf{55.56 $\pm$ 0.00} & 81.95 $\pm$ 0.19 & \textbf{65.42 $\pm$ 0.76} & 33.33 $\pm$ 0.00 & 54.40 $\pm$ 2.07 & \underline{4.22} \\
 & \emph{RepNodes (best)} & 25.39 $\pm$ 1.08 & - & - & 83.33 $\pm$ 0.00 & - & - & - & \textbf{43.33 $\pm$ 2.34} & - & \textbf{3.00} \\
 & \emph{RepEdges (SRL*)} & 25.11 $\pm$ 0.65 & \underline{71.72 $\pm$ 0.28} & 69.04 $\pm$ 0.13 & 83.41 $\pm$ 0.16 & \underline{50.00 $\pm$ 0.00} & 81.91 $\pm$ 0.30 & \underline{63.50 $\pm$ 0.72} & 33.33 $\pm$ 0.00 & 48.80 $\pm$ 1.69 & 6.33 \\
 & \emph{RepEdges (best)} & 25.39 $\pm$ 1.08 & - & 69.79 $\pm$ 0.29 & - & - & 81.93 $\pm$ 0.23 & - & 36.67 $\pm$ 4.68 & 49.60 $\pm$ 3.37 & 4.89 \\
\cmidrule{1-11} \cmidrule(l){12-12}
\multicolumn{12}{l}{\textbf{GAT}} \\
 & Baseline & 24.97 $\pm$ 0.32 & 66.34 $\pm$ 2.03 & 61.02 $\pm$ 0.74 & \underline{75.48 $\pm$ 1.38} & \underline{43.33 $\pm$ 2.34} & \underline{77.73 $\pm$ 0.91} & 45.77 $\pm$ 2.24 & 36.67 $\pm$ 4.68 & 51.20 $\pm$ 4.92 & 6.56 \\
 & BORF & 25.47 $\pm$ 0.76 & \underline{67.67 $\pm$ 2.62} & \underline{62.95 $\pm$ 1.35} & 72.67 $\pm$ 3.96 & 42.22 $\pm$ 2.72 & OOM & OOM & 36.67 $\pm$ 4.44 & \textbf{56.00 $\pm$ 6.69} & 6.78 \\
 & FOSR & 26.03 $\pm$ 0.62 & 66.26 $\pm$ 1.41 & 61.39 $\pm$ 3.14 & 73.70 $\pm$ 4.42 & \textbf{44.44 $\pm$ 0.00} & 76.94 $\pm$ 0.94 & 45.58 $\pm$ 1.00 & 35.56 $\pm$ 5.67 & \underline{55.20 $\pm$ 4.66} & 5.33 \\
 & SDRF & 25.95 $\pm$ 0.68 & 65.99 $\pm$ 1.43 & 61.02 $\pm$ 0.70 & 73.70 $\pm$ 1.42 & \underline{43.33 $\pm$ 2.22} & OOM & 45.85 $\pm$ 1.25 & 34.44 $\pm$ 2.22 & 53.60 $\pm$ 3.20 & 7.78 \\
 & JDR & 25.34 $\pm$ 0.41 & 40.97 $\pm$ 2.10 & 37.53 $\pm$ 1.09 & 46.30 $\pm$ 3.39 & \textbf{44.44 $\pm$ 0.00} & \textbf{82.43 $\pm$ 0.48} & \textbf{51.88 $\pm$ 2.33} & \textbf{57.78 $\pm$ 2.72} & 52.00 $\pm$ 0.00 & 6.33 \\
 & ComFy & \textbf{28.21 $\pm$ 0.54} & 65.55 $\pm$ 2.56 & \textbf{69.52 $\pm$ 3.36} & \textbf{78.96 $\pm$ 1.47} & 30.00 $\pm$ 8.43 & 77.28 $\pm$ 1.70 & 46.19 $\pm$ 1.76 & 27.78 $\pm$ 0.00 & 48.00 $\pm$ 0.00 & 6.33 \\
 & TRIGON & \underline{27.47 $\pm$ 1.23} & 45.90 $\pm$ 10.27 & 47.71 $\pm$ 3.89 & 55.19 $\pm$ 10.23 & 28.89 $\pm$ 22.01 & 49.30 $\pm$ 1.84 & 45.15 $\pm$ 2.38 & 30.00 $\pm$ 6.33 & 48.00 $\pm$ 6.32 & 9.78 \\
 & MN & 25.29 $\pm$ 0.58 & 67.31 $\pm$ 2.11 & OOM & 73.85 $\pm$ 1.45 & \textbf{44.44 $\pm$ 0.00} & OOM & \underline{47.62 $\pm$ 2.87} & 36.67 $\pm$ 5.81 & 50.40 $\pm$ 4.92 & 6.56 \\[0.15cm]
 & \emph{RepNodes (SRL*)} & 26.00 $\pm$ 0.59 & \textbf{68.81 $\pm$ 0.80} & 59.94 $\pm$ 1.62 & 70.30 $\pm$ 5.37 & \textbf{44.44 $\pm$ 0.00} & 75.83 $\pm$ 0.44 & 46.77 $\pm$ 3.28 & 33.33 $\pm$ 0.00 & \textbf{56.00 $\pm$ 2.67} & 5.11 \\
 & \emph{RepNodes (best)} & - & - & 61.20 $\pm$ 1.67 & 73.85 $\pm$ 1.49 & - & 76.56 $\pm$ 1.01 & \underline{47.62 $\pm$ 2.94} & \underline{37.78 $\pm$ 4.38} & - & \textbf{2.56} \\
 & \emph{RepEdges (SRL*)} & 26.00 $\pm$ 0.59 & \textbf{68.81 $\pm$ 0.80} & 59.94 $\pm$ 1.62 & 70.30 $\pm$ 5.37 & \textbf{44.44 $\pm$ 0.00} & 75.89 $\pm$ 0.49 & 46.77 $\pm$ 3.28 & 33.33 $\pm$ 0.00 & \textbf{56.00 $\pm$ 2.67} & 5.00 \\
 & \emph{RepEdges (best)} & - & - & 61.14 $\pm$ 1.70 & 73.85 $\pm$ 1.49 & - & 76.56 $\pm$ 1.01 & \underline{47.62 $\pm$ 2.95} & \underline{37.78 $\pm$ 4.38} & - & \underline{2.67} \\
\cmidrule{1-11} \cmidrule(l){12-12}
\multicolumn{12}{l}{\textbf{GIN}} \\
 & Baseline & 24.66 $\pm$ 0.96 & 63.35 $\pm$ 0.68 & 68.67 $\pm$ 0.72 & 82.89 $\pm$ 0.80 & 47.78 $\pm$ 5.97 & 80.43 $\pm$ 0.43 & 64.65 $\pm$ 4.07 & \underline{54.44 $\pm$ 2.34} & 48.00 $\pm$ 2.67 & 7.11 \\
 & BORF & 25.00 $\pm$ 0.67 & 64.32 $\pm$ 3.49 & \underline{68.86 $\pm$ 0.86} & 82.15 $\pm$ 1.08 & \underline{52.22 $\pm$ 2.72} & OOM & OOM & 52.22 $\pm$ 2.72 & \textbf{60.80 $\pm$ 5.88} & 6.33 \\
 & FOSR & 24.92 $\pm$ 1.40 & 63.17 $\pm$ 5.92 & 68.07 $\pm$ 1.45 & 81.93 $\pm$ 1.00 & 46.67 $\pm$ 2.72 & 80.07 $\pm$ 0.36 & \underline{65.38 $\pm$ 1.71} & 34.44 $\pm$ 2.22 & \underline{59.20 $\pm$ 4.66} & 7.33 \\
 & SDRF & 24.92 $\pm$ 1.38 & 63.26 $\pm$ 5.17 & 68.67 $\pm$ 0.69 & \underline{82.96 $\pm$ 0.47} & 48.89 $\pm$ 4.16 & OOM & 65.35 $\pm$ 1.67 & 47.78 $\pm$ 2.72 & 51.20 $\pm$ 2.99 & 6.33 \\
 & JDR & 25.03 $\pm$ 1.07 & 52.51 $\pm$ 5.13 & 52.29 $\pm$ 0.93 & 68.37 $\pm$ 0.90 & 48.89 $\pm$ 8.89 & \textbf{82.58 $\pm$ 0.65} & 27.73 $\pm$ 1.61 & 46.67 $\pm$ 2.72 & 53.60 $\pm$ 3.20 & 7.78 \\
 & ComFy & \underline{26.08 $\pm$ 0.99} & \underline{66.26 $\pm$ 3.98} & \textbf{72.23 $\pm$ 0.69} & \textbf{85.33 $\pm$ 0.62} & 27.78 $\pm$ 10.39 & \underline{82.02 $\pm$ 0.20} & 58.08 $\pm$ 2.65 & 27.78 $\pm$ 0.00 & 56.80 $\pm$ 7.16 & 5.22 \\
 & TRIGON & \textbf{26.89 $\pm$ 2.38} & 41.23 $\pm$ 19.00 & 55.30 $\pm$ 4.37 & 69.70 $\pm$ 8.50 & 45.56 $\pm$ 8.24 & 53.97 $\pm$ 3.08 & 59.42 $\pm$ 1.30 & 27.78 $\pm$ 0.00 & 55.20 $\pm$ 3.35 & 8.67 \\
 & MN & 25.13 $\pm$ 0.91 & 61.94 $\pm$ 4.53 & OOM & 82.59 $\pm$ 0.80 & 45.56 $\pm$ 2.28 & OOM & 64.73 $\pm$ 3.47 & \textbf{58.89 $\pm$ 2.79} & 49.60 $\pm$ 3.28 & 8.00 \\[0.15cm]
 & \emph{RepNodes (SRL*)} & 25.11 $\pm$ 0.58 & \textbf{71.89 $\pm$ 2.76} & 68.01 $\pm$ 1.60 & \underline{82.96 $\pm$ 0.65} & 46.67 $\pm$ 2.87 & 80.42 $\pm$ 0.53 & \textbf{67.27 $\pm$ 0.94} & 41.11 $\pm$ 4.68 & 51.20 $\pm$ 4.13 & 5.22 \\
 & \emph{RepNodes (best)} & 25.13 $\pm$ 0.93 & - & 68.55 $\pm$ 1.51 & - & \textbf{55.56 $\pm$ 0.00} & 80.49 $\pm$ 0.58 & - & \textbf{58.89 $\pm$ 2.87} & 52.00 $\pm$ 4.62 & \textbf{2.56} \\
 & \emph{RepEdges (SRL*)} & 25.11 $\pm$ 0.58 & \textbf{71.89 $\pm$ 2.76} & 68.01 $\pm$ 1.60 & \underline{82.96 $\pm$ 0.65} & 46.67 $\pm$ 2.87 & 80.42 $\pm$ 0.53 & \underline{65.38 $\pm$ 2.29} & 41.11 $\pm$ 4.68 & 51.20 $\pm$ 4.13 & 5.44 \\
 & \emph{RepEdges (best)} & 25.13 $\pm$ 0.93 & - & 68.55 $\pm$ 1.51 & - & \textbf{55.56 $\pm$ 0.00} & 80.49 $\pm$ 0.58 & - & \textbf{58.89 $\pm$ 2.87} & 52.00 $\pm$ 4.62 & \underline{2.78} \\
\bottomrule
\end{tabular}}
\caption{Node classification test accuracies at L=2 on homophilic/heterophilic datasets. For RAwR, we report both the result for the epsilon chosen by a grid search on validation set and the one chosen by the SRL* heuristic; if they coincide we report '-'. OOM stands for Out Of Memory. Best per dataset is bold, second-best is underlined.}
\label{tab:all_models_rewiring_comparison}
\end{table}

\begin{table}[H]
\centering
\resizebox{0.9\textwidth}{!}{%
\begin{tabular}{llcccccc c}
\toprule
 &  & Caterpillar & Grid & Ladder & Line & Lobster & Tree & Avg.Rank \\
\cmidrule{1-8} \cmidrule(l){9-9}
\multicolumn{9}{c}{\emph{WITH NODE FEATURES}}
\\
\cmidrule{1-8} \cmidrule(l){9-9}
\multicolumn{9}{l}{\textbf{GCN}} \\
 & Baseline & 41.18 $\pm$ 0.00 & 71.82 $\pm$ 2.59 & 60.00 $\pm$ 6.39 & \textbf{100.00 $\pm$ 0.00} & 78.33 $\pm$ 7.45 & \textbf{100.00 $\pm$ 0.00} & 5.17 \\
 & BORF & 43.53 $\pm$ 3.22 & 71.82 $\pm$ 2.03 & 54.29 $\pm$ 3.91 & 55.00 $\pm$ 32.60 & 83.33 $\pm$ 0.00 & \textbf{100.00 $\pm$ 0.00} & 6.33 \\
 & FOSR & 40.00 $\pm$ 2.63 & 62.73 $\pm$ 2.03 & 58.57 $\pm$ 5.98 & 50.00 $\pm$ 0.00 & 78.33 $\pm$ 7.45 & \underline{82.50 $\pm$ 6.85} & 8.83 \\
 & SDRF & 42.35 $\pm$ 2.63 & \underline{74.09 $\pm$ 2.59} & \textbf{64.29 $\pm$ 5.05} & \textbf{100.00 $\pm$ 0.00} & 78.33 $\pm$ 7.45 & \textbf{100.00 $\pm$ 0.00} & 4.00 \\
 & JDR & 11.76 $\pm$ 0.00 & 10.45 $\pm$ 3.69 & 37.14 $\pm$ 7.00 & 20.00 $\pm$ 10.00 & 50.00 $\pm$ 0.00 & 15.00 $\pm$ 14.58 & 11.83 \\
 & ComFy & 51.76 $\pm$ 6.44 & 58.18 $\pm$ 6.93 & \underline{62.86 $\pm$ 9.31} & \underline{70.00 $\pm$ 11.18} & 83.33 $\pm$ 0.00 & 42.50 $\pm$ 25.92 & 6.67 \\
 & TRIGON & 68.24 $\pm$ 3.22 & 41.82 $\pm$ 5.23 & 45.71 $\pm$ 11.95 & 20.00 $\pm$ 20.92 & \textbf{95.00 $\pm$ 4.56} & 20.00 $\pm$ 18.96 & 8.33 \\
 & MN & 44.71 $\pm$ 3.22 & \textbf{76.36 $\pm$ 2.59} & \underline{62.86 $\pm$ 3.19} & \textbf{100.00 $\pm$ 0.00} & \underline{91.67 $\pm$ 0.00} & \textbf{100.00 $\pm$ 0.00} & 2.33 \\[0.15cm]
 & \emph{RepNodes (SRL*)} & \textbf{76.47 $\pm$ 0.00} & 57.27 $\pm$ 1.90 & 55.71 $\pm$ 9.31 & \textbf{100.00 $\pm$ 0.00} & 75.00 $\pm$ 0.00 & \textbf{100.00 $\pm$ 0.00} & 5.50 \\
 & \emph{RepNodes (best)} & - & \textbf{76.36 $\pm$ 2.59} & \underline{62.86 $\pm$ 3.19} & - & \underline{91.67 $\pm$ 0.00} & - & \textbf{1.33} \\
 & \emph{RepEdges (SRL*)} & \underline{70.59 $\pm$ 0.00} & 68.64 $\pm$ 1.90 & 57.14 $\pm$ 8.75 & \textbf{100.00 $\pm$ 0.00} & 83.33 $\pm$ 0.00 & \textbf{100.00 $\pm$ 0.00} & 4.17 \\
 & \emph{RepEdges (best)} & - & \textbf{76.36 $\pm$ 2.59} & \underline{62.86 $\pm$ 3.19} & - & \underline{91.67 $\pm$ 0.00} & - & \underline{1.67} \\
\cmidrule{1-8} \cmidrule(l){9-9}
\multicolumn{9}{l}{\textbf{GAT}} \\
 & Baseline & 68.24 $\pm$ 20.63 & \underline{58.64 $\pm$ 4.93} & \underline{61.43 $\pm$ 9.58} & 70.00 $\pm$ 11.18 & 73.33 $\pm$ 3.73 & \underline{77.50 $\pm$ 10.46} & 5.00 \\
 & BORF & 60.00 $\pm$ 17.35 & 53.64 $\pm$ 5.23 & 57.14 $\pm$ 7.14 & 70.00 $\pm$ 27.39 & \underline{86.67 $\pm$ 4.56} & 67.50 $\pm$ 18.96 & 7.00 \\
 & FOSR & 70.59 $\pm$ 8.32 & 56.36 $\pm$ 5.43 & 57.14 $\pm$ 11.29 & 25.00 $\pm$ 25.00 & 66.67 $\pm$ 5.89 & 52.50 $\pm$ 16.30 & 8.00 \\
 & SDRF & \textbf{81.18 $\pm$ 8.72} & \textbf{62.73 $\pm$ 2.59} & 58.57 $\pm$ 14.64 & 70.00 $\pm$ 11.18 & 73.33 $\pm$ 3.73 & \textbf{85.00 $\pm$ 10.46} & \underline{3.83} \\
 & JDR & 11.76 $\pm$ 0.00 & 15.91 $\pm$ 0.00 & 31.43 $\pm$ 5.71 & 0.00 $\pm$ 0.00 & 50.00 $\pm$ 0.00 & 0.00 $\pm$ 0.00 & 12.00 \\
 & ComFy & 57.65 $\pm$ 4.92 & 56.36 $\pm$ 2.49 & \textbf{62.86 $\pm$ 9.31} & 70.00 $\pm$ 11.18 & \textbf{88.33 $\pm$ 4.56} & 42.50 $\pm$ 25.92 & 5.33 \\
 & TRIGON & 71.76 $\pm$ 2.63 & 41.82 $\pm$ 6.14 & 41.43 $\pm$ 15.49 & 30.00 $\pm$ 20.92 & 83.33 $\pm$ 10.21 & 15.00 $\pm$ 10.46 & 8.33 \\
 & MN & 71.76 $\pm$ 28.03 & 55.91 $\pm$ 5.47 & 52.86 $\pm$ 13.92 & \underline{95.00 $\pm$ 11.18} & 78.33 $\pm$ 4.56 & \underline{77.50 $\pm$ 10.46} & 5.67 \\[0.15cm]
 & \emph{RepNodes (SRL*)} & 64.71 $\pm$ 23.53 & 46.36 $\pm$ 3.80 & 60.00 $\pm$ 13.92 & 75.00 $\pm$ 17.68 & 73.33 $\pm$ 16.03 & \textbf{85.00 $\pm$ 16.30} & 5.33 \\
 & \emph{RepNodes (best)} & \underline{78.82 $\pm$ 22.24} & 55.91 $\pm$ 5.47 & - & \textbf{100.00 $\pm$ 0.00} & 80.00 $\pm$ 9.50 & - & \textbf{2.67} \\
 & \emph{RepEdges (SRL*)} & 64.71 $\pm$ 23.53 & 46.36 $\pm$ 3.80 & 60.00 $\pm$ 13.92 & 75.00 $\pm$ 17.68 & 73.33 $\pm$ 16.03 & \textbf{85.00 $\pm$ 16.30} & 5.33 \\
 & \emph{RepEdges (best)} & \underline{78.82 $\pm$ 22.24} & 55.91 $\pm$ 5.47 & - & \textbf{100.00 $\pm$ 0.00} & 80.00 $\pm$ 9.50 & - & \textbf{2.67} \\
\cmidrule{1-8} \cmidrule(l){9-9}
\multicolumn{9}{l}{\textbf{GIN}} \\
 & Baseline & 49.41 $\pm$ 8.92 & 71.36 $\pm$ 8.59 & 54.29 $\pm$ 10.83 & \textbf{100.00 $\pm$ 0.00} & 76.67 $\pm$ 3.73 & \textbf{100.00 $\pm$ 0.00} & 5.83 \\
 & BORF & 67.06 $\pm$ 10.69 & 70.45 $\pm$ 10.16 & 60.00 $\pm$ 10.83 & \underline{65.00 $\pm$ 37.91} & 73.33 $\pm$ 3.73 & \underline{97.50 $\pm$ 5.59} & 7.33 \\
 & FOSR & 49.41 $\pm$ 15.34 & 60.91 $\pm$ 4.93 & 54.29 $\pm$ 6.39 & 45.00 $\pm$ 32.60 & 78.33 $\pm$ 11.18 & 77.50 $\pm$ 10.46 & 9.17 \\
 & SDRF & 56.47 $\pm$ 18.88 & \textbf{75.91 $\pm$ 5.93} & 55.71 $\pm$ 11.74 & \textbf{100.00 $\pm$ 0.00} & 76.67 $\pm$ 3.73 & 95.00 $\pm$ 6.85 & 5.83 \\
 & JDR & 11.76 $\pm$ 0.00 & 12.73 $\pm$ 2.32 & 52.86 $\pm$ 12.45 & 0.00 $\pm$ 0.00 & 41.67 $\pm$ 7.45 & 10.00 $\pm$ 5.00 & 11.83 \\
 & ComFy & 55.29 $\pm$ 5.26 & 60.91 $\pm$ 6.31 & \textbf{62.86 $\pm$ 11.74} & 60.00 $\pm$ 13.69 & 88.33 $\pm$ 12.64 & 50.00 $\pm$ 17.68 & 7.33 \\
 & TRIGON & \underline{68.24 $\pm$ 6.71} & 40.00 $\pm$ 2.03 & 41.43 $\pm$ 5.98 & 35.00 $\pm$ 13.69 & \textbf{91.67 $\pm$ 0.00} & 22.50 $\pm$ 13.69 & 8.50 \\
 & MN & 58.82 $\pm$ 17.15 & \underline{72.27 $\pm$ 4.37} & 60.00 $\pm$ 8.14 & \textbf{100.00 $\pm$ 0.00} & 80.00 $\pm$ 4.56 & \textbf{100.00 $\pm$ 0.00} & 4.00 \\[0.15cm]
 & \emph{RepNodes (SRL*)} & \textbf{77.65 $\pm$ 4.92} & 63.64 $\pm$ 6.63 & \underline{61.43 $\pm$ 6.39} & \textbf{100.00 $\pm$ 0.00} & \underline{90.00 $\pm$ 3.73} & \textbf{100.00 $\pm$ 0.00} & \underline{2.33} \\
 & \emph{RepNodes (best)} & - & \underline{72.27 $\pm$ 4.37} & - & - & - & - & \textbf{1.50} \\
 & \emph{RepEdges (SRL*)} & \textbf{77.65 $\pm$ 4.92} & 63.64 $\pm$ 6.63 & \underline{61.43 $\pm$ 6.39} & \textbf{100.00 $\pm$ 0.00} & \underline{90.00 $\pm$ 3.73} & \textbf{100.00 $\pm$ 0.00} & \underline{2.33} \\
 & \emph{RepEdges (best)} & - & \underline{72.27 $\pm$ 4.37} & - & - & - & - & \textbf{1.50} \\
\cmidrule{1-8} \cmidrule(l){9-9}
\multicolumn{9}{c}{\emph{WITHOUT NODE FEATURES}}
\\
\cmidrule{1-8} \cmidrule(l){9-9}
\multicolumn{9}{l}{\textbf{GCN}} \\
 & Baseline & 38.82 $\pm$ 3.22 & 72.73 $\pm$ 4.25 & \underline{62.86 $\pm$ 3.19} & \textbf{100.00 $\pm$ 0.00} & 76.67 $\pm$ 3.73 & \textbf{100.00 $\pm$ 0.00} & 4.17 \\
 & BORF & 38.82 $\pm$ 5.26 & 72.27 $\pm$ 5.18 & 57.14 $\pm$ 5.05 & 85.00 $\pm$ 33.54 & \underline{85.00 $\pm$ 3.73} & \textbf{100.00 $\pm$ 0.00} & 6.50 \\
 & FOSR & 36.47 $\pm$ 7.67 & 66.36 $\pm$ 1.02 & 55.71 $\pm$ 3.19 & 50.00 $\pm$ 17.68 & 73.33 $\pm$ 3.73 & \underline{87.50 $\pm$ 8.84} & 10.33 \\
 & SDRF & 38.82 $\pm$ 3.22 & \underline{73.64 $\pm$ 2.03} & 58.57 $\pm$ 9.31 & \textbf{100.00 $\pm$ 0.00} & 76.67 $\pm$ 3.73 & \textbf{100.00 $\pm$ 0.00} & 5.17 \\
 & JDR & 44.71 $\pm$ 2.88 & 60.00 $\pm$ 1.11 & 54.29 $\pm$ 7.28 & \underline{95.00 $\pm$ 10.00} & 73.33 $\pm$ 3.33 & \textbf{100.00 $\pm$ 0.00} & 8.33 \\
 & ComFy & 64.71 $\pm$ 4.16 & 61.36 $\pm$ 3.94 & \underline{62.86 $\pm$ 5.98} & 75.00 $\pm$ 30.62 & 78.33 $\pm$ 4.56 & 50.00 $\pm$ 12.50 & 7.33 \\
 & TRIGON & 67.06 $\pm$ 24.47 & 59.09 $\pm$ 12.96 & \textbf{65.71 $\pm$ 5.98} & 35.00 $\pm$ 22.36 & 20.00 $\pm$ 27.39 & 50.00 $\pm$ 0.00 & 8.83 \\
 & MN & 45.88 $\pm$ 2.63 & \textbf{75.45 $\pm$ 1.90} & 61.43 $\pm$ 3.91 & \textbf{100.00 $\pm$ 0.00} & \textbf{91.67 $\pm$ 0.00} & \textbf{100.00 $\pm$ 0.00} & 2.50 \\[0.15cm]
 & \emph{RepNodes (SRL*)} & \textbf{76.47 $\pm$ 0.00} & 60.00 $\pm$ 2.03 & 60.00 $\pm$ 3.91 & \textbf{100.00 $\pm$ 0.00} & 75.00 $\pm$ 0.00 & \textbf{100.00 $\pm$ 0.00} & 5.00 \\
 & \emph{RepNodes (best)} & - & \textbf{75.45 $\pm$ 1.90} & 61.43 $\pm$ 3.91 & - & \textbf{91.67 $\pm$ 0.00} & - & \textbf{1.50} \\
 & \emph{RepEdges (SRL*)} & \underline{70.59 $\pm$ 0.00} & 71.82 $\pm$ 3.45 & 61.43 $\pm$ 3.91 & \textbf{100.00 $\pm$ 0.00} & 83.33 $\pm$ 0.00 & \textbf{100.00 $\pm$ 0.00} & 3.50 \\
 & \emph{RepEdges (best)} & - & \textbf{75.45 $\pm$ 1.90} & - & - & \textbf{91.67 $\pm$ 0.00} & - & \underline{1.83} \\
\cmidrule{1-8} \cmidrule(l){9-9}
\multicolumn{9}{l}{\textbf{GAT}} \\
 & Baseline & \underline{69.41 $\pm$ 18.32} & \underline{56.82 $\pm$ 4.25} & 52.86 $\pm$ 8.14 & \underline{95.00 $\pm$ 11.18} & 76.67 $\pm$ 9.13 & 85.00 $\pm$ 5.59 & 5.33 \\
 & BORF & 57.65 $\pm$ 20.96 & 55.91 $\pm$ 3.45 & 54.29 $\pm$ 12.98 & 75.00 $\pm$ 17.68 & \underline{85.00 $\pm$ 6.97} & 85.00 $\pm$ 10.46 & 6.83 \\
 & FOSR & 43.53 $\pm$ 6.71 & 53.18 $\pm$ 4.43 & 55.71 $\pm$ 5.98 & 30.00 $\pm$ 32.60 & 56.67 $\pm$ 3.73 & 65.00 $\pm$ 13.69 & 9.50 \\
 & SDRF & 60.00 $\pm$ 17.84 & \textbf{58.18 $\pm$ 5.70} & 58.57 $\pm$ 7.82 & \underline{95.00 $\pm$ 11.18} & 76.67 $\pm$ 9.13 & 80.00 $\pm$ 6.85 & 5.33 \\
 & JDR & 14.12 $\pm$ 4.71 & 26.82 $\pm$ 4.85 & 35.71 $\pm$ 0.00 & 0.00 $\pm$ 0.00 & 53.33 $\pm$ 6.67 & 0.00 $\pm$ 0.00 & 12.00 \\
 & ComFy & 68.24 $\pm$ 5.26 & 56.36 $\pm$ 2.49 & \underline{60.00 $\pm$ 6.39} & 80.00 $\pm$ 27.39 & \textbf{91.67 $\pm$ 0.00} & 55.00 $\pm$ 11.18 & 5.33 \\
 & TRIGON & 60.00 $\pm$ 20.96 & 53.18 $\pm$ 10.11 & \underline{60.00 $\pm$ 9.58} & 55.00 $\pm$ 27.39 & 75.00 $\pm$ 13.18 & 62.50 $\pm$ 19.76 & 8.00 \\
 & MN & 51.76 $\pm$ 20.55 & 56.36 $\pm$ 5.66 & \textbf{64.29 $\pm$ 12.37} & 80.00 $\pm$ 27.39 & 75.00 $\pm$ 0.00 & \underline{87.50 $\pm$ 8.84} & 5.83 \\[0.15cm]
 & \emph{RepNodes (SRL*)} & \textbf{85.88 $\pm$ 19.77} & 50.91 $\pm$ 2.59 & 50.00 $\pm$ 8.75 & 90.00 $\pm$ 13.69 & 81.67 $\pm$ 9.13 & \textbf{90.00 $\pm$ 10.46} & \underline{5.00} \\
 & \emph{RepNodes (best)} & - & 56.36 $\pm$ 5.66 & \textbf{64.29 $\pm$ 12.37} & \textbf{100.00 $\pm$ 0.00} & - & - & \textbf{1.67} \\
 & \emph{RepEdges (SRL*)} & \textbf{85.88 $\pm$ 19.77} & 50.91 $\pm$ 2.59 & 50.00 $\pm$ 8.75 & 90.00 $\pm$ 13.69 & 81.67 $\pm$ 9.13 & \textbf{90.00 $\pm$ 10.46} & \underline{5.00} \\
 & \emph{RepEdges (best)} & - & 56.36 $\pm$ 5.66 & \textbf{64.29 $\pm$ 12.37} & \textbf{100.00 $\pm$ 0.00} & - & - & \textbf{1.67} \\
\cmidrule{1-8} \cmidrule(l){9-9}
\multicolumn{9}{l}{\textbf{GIN}} \\
 & Baseline & 63.53 $\pm$ 15.23 & \underline{74.55 $\pm$ 4.37} & 55.71 $\pm$ 10.59 & \textbf{100.00 $\pm$ 0.00} & \underline{81.67 $\pm$ 6.97} & \textbf{100.00 $\pm$ 0.00} & 4.83 \\
 & BORF & \underline{70.59 $\pm$ 16.64} & 70.00 $\pm$ 2.49 & \underline{61.43 $\pm$ 10.83} & 60.00 $\pm$ 54.77 & 75.00 $\pm$ 0.00 & \textbf{100.00 $\pm$ 0.00} & 5.67 \\
 & FOSR & 60.00 $\pm$ 17.84 & 68.64 $\pm$ 2.96 & 57.14 $\pm$ 5.05 & 20.00 $\pm$ 20.92 & 80.00 $\pm$ 9.50 & \underline{77.50 $\pm$ 10.46} & 9.17 \\
 & SDRF & 63.53 $\pm$ 7.67 & 73.18 $\pm$ 2.96 & \underline{61.43 $\pm$ 3.91} & \textbf{100.00 $\pm$ 0.00} & \underline{81.67 $\pm$ 6.97} & \textbf{100.00 $\pm$ 0.00} & 3.50 \\
 & JDR & 42.35 $\pm$ 6.86 & 31.82 $\pm$ 3.52 & 42.86 $\pm$ 4.52 & 5.00 $\pm$ 10.00 & 70.00 $\pm$ 8.50 & 25.00 $\pm$ 7.91 & 11.83 \\
 & ComFy & 69.41 $\pm$ 4.92 & 63.18 $\pm$ 4.66 & \underline{61.43 $\pm$ 15.65} & \underline{70.00 $\pm$ 32.60} & 80.00 $\pm$ 7.45 & 57.50 $\pm$ 11.18 & 7.50 \\
 & TRIGON & 57.65 $\pm$ 15.78 & 25.00 $\pm$ 29.68 & \textbf{64.29 $\pm$ 12.37} & 25.00 $\pm$ 30.62 & \underline{81.67 $\pm$ 23.12} & 72.50 $\pm$ 18.54 & 8.17 \\
 & MN & 61.18 $\pm$ 12.20 & \textbf{75.45 $\pm$ 8.56} & \underline{61.43 $\pm$ 8.14} & \textbf{100.00 $\pm$ 0.00} & 76.67 $\pm$ 3.73 & \textbf{100.00 $\pm$ 0.00} & 4.00 \\[0.15cm]
 & \emph{RepNodes (SRL*)} & \textbf{77.65 $\pm$ 2.63} & 67.73 $\pm$ 2.96 & 60.00 $\pm$ 8.14 & \textbf{100.00 $\pm$ 0.00} & \textbf{88.33 $\pm$ 4.56} & \textbf{100.00 $\pm$ 0.00} & \underline{3.33} \\
 & \emph{RepNodes (best)} & - & \textbf{75.45 $\pm$ 8.56} & \underline{61.43 $\pm$ 8.14} & - & - & - & \textbf{1.17} \\
 & \emph{RepEdges (SRL*)} & \textbf{77.65 $\pm$ 2.63} & 67.73 $\pm$ 2.96 & 60.00 $\pm$ 8.14 & \textbf{100.00 $\pm$ 0.00} & \textbf{88.33 $\pm$ 4.56} & \textbf{100.00 $\pm$ 0.00} & \underline{3.33} \\
 & \emph{RepEdges (best)} & - & \textbf{75.45 $\pm$ 8.56} & \underline{61.43 $\pm$ 8.14} & - & - & - & \textbf{1.17} \\
\bottomrule
\end{tabular}}
\caption{Node classification test accuracies at L=2 on CanYouHearMeNow datasets. For RAwR, we report both the result for the epsilon chosen by a grid search on validation set and the one chosen by the SRL* heuristic; if they coincide we report '-'. OOM stands for Out Of Memory. Best per dataset is bold, second-best is underlined.}
\label{tab:all_models_cyhmn_rewiring_comparison}
\end{table}

\begin{table*}[h]
\centering
\resizebox{\textwidth}{!}{%
\begin{tabular}{llllllllllll}
\toprule
 &  & Dataset & Actor & Chameleon & Citeseer & Cora & Cornell & PubMed & Squirrel & Texas & Wisconsin \\
Model & Features & Method &  &  &  &  &  &  &  &  &  \\
\midrule
\multirow[t]{8}{*}{GCN} & \multirow[t]{4}{*}{Yes} & Baseline  & 25.74 $\pm$ 0.93 & 65.02 $\pm$ 0.59 & 75.66 $\pm$ 0.65 & 84.07 $\pm$ 0.94 & 48.89 $\pm$ 2.48 & 87.63 $\pm$ 0.24 & 61.65 $\pm$ 0.22 & \textbf{44.44 $\pm$ 3.93} & 58.40 $\pm$ 3.58 \\
 &  & MN  & \underline{25.97 $\pm$ 0.88} & \underline{66.30 $\pm$ 0.84} & \multicolumn{1}{c}{OOM} & \underline{84.96 $\pm$ 0.63} & 44.44 $\pm$ 0.00 & \multicolumn{1}{c}{OOM} & 61.17 $\pm$ 0.29 & 38.89 $\pm$ 3.70 & 53.60 $\pm$ 5.72 \\
 &  & RepNodes  & \cellcolor{green!30}\textbf{26.45 $\pm$ 0.71} & \cellcolor{green!30}\textbf{73.83 $\pm$ 0.50} & \cellcolor{green!30}\textbf{76.02 $\pm$ 1.08} & \cellcolor{green!30}\textbf{85.26 $\pm$ 0.61} & 46.67 $\pm$ 3.04 & \cellcolor{green!30}\textbf{87.77 $\pm$ 0.21} & \cellcolor{green!30}\textbf{65.77 $\pm$ 0.38} & 38.89 $\pm$ 3.93 & \cellcolor{green!30}\textbf{64.80 $\pm$ 1.79} \\
 &  & RepEdges  & \cellcolor{green!30}{26.21 $\pm$ 0.80} & \cellcolor{green!30}{71.37 $\pm$ 0.31} & \cellcolor{green!30}{76.02 $\pm$ 1.20} & \cellcolor{green!30}{84.96 $\pm$ 0.77} & \cellcolor{green!30}{52.22 $\pm$ 4.97} & \cellcolor{green!30}{87.75 $\pm$ 0.19} & \cellcolor{green!30}{63.46 $\pm$ 0.19} & 38.89 $\pm$ 3.93 & \cellcolor{green!30}{60.80 $\pm$ 9.96} \\
\cline{2-12}
 & \multirow[t]{4}{*}{No} & Baseline  & \textbf{25.45 $\pm$ 0.78} & 67.84 $\pm$ 0.93 & \textbf{69.73 $\pm$ 0.57} & \textbf{83.48 $\pm$ 0.19} & 46.67 $\pm$ 2.87 & \textbf{82.12 $\pm$ 0.06} & 60.85 $\pm$ 0.26 & \textbf{47.78 $\pm$ 7.94} & 54.40 $\pm$ 3.37 \\
 &  & MN  & 25.39 $\pm$ 0.90 & \underline{68.24 $\pm$ 0.35} & \multicolumn{1}{c}{OOM} & 83.24 $\pm$ 0.32 & 44.44 $\pm$ 0.00 & \multicolumn{1}{c}{OOM} & 60.73 $\pm$ 0.53 & 40.00 $\pm$ 5.58 & 53.60 $\pm$ 3.28 \\
 &  & RepNodes  & 25.39 $\pm$ 1.08 & \cellcolor{green!30}\textbf{75.07 $\pm$ 0.23} & 69.46 $\pm$ 0.93 & 83.33 $\pm$ 0.00 & \cellcolor{green!30}{48.89 $\pm$ 4.38} & 81.95 $\pm$ 0.19 & \cellcolor{green!30}\textbf{65.42 $\pm$ 0.76} & 41.11 $\pm$ 2.87 & \cellcolor{green!30}\textbf{59.60 $\pm$ 6.65} \\
 &  & RepEdges  & 25.39 $\pm$ 1.08 & \cellcolor{green!30}{71.72 $\pm$ 0.28} & 69.52 $\pm$ 0.71 & 83.41 $\pm$ 0.16 & \cellcolor{green!30}\textbf{50.00 $\pm$ 5.24} & 81.93 $\pm$ 0.23 & \cellcolor{green!30}{63.50 $\pm$ 0.72} & 43.33 $\pm$ 10.08 & \cellcolor{green!30}{59.20 $\pm$ 8.18} \\
\cline{1-12} \cline{2-12}
\midrule
\multirow[t]{8}{*}{GAT} & \multirow[t]{4}{*}{Yes} & Baseline  & \textbf{25.39 $\pm$ 1.02} & 60.09 $\pm$ 5.47 & 74.04 $\pm$ 0.39 & 84.89 $\pm$ 0.61 & 46.67 $\pm$ 3.04 & 86.67 $\pm$ 0.35 & 43.50 $\pm$ 1.55 & \textbf{48.89 $\pm$ 2.48} & 55.20 $\pm$ 5.22 \\
 &  & MN  & 23.39 $\pm$ 1.28 & 60.00 $\pm$ 5.20 & \multicolumn{1}{c}{OOM} & 83.56 $\pm$ 0.72 & 42.22 $\pm$ 2.87 & \multicolumn{1}{c}{OOM} & 41.96 $\pm$ 2.75 & 35.56 $\pm$ 4.68 & \underline{56.80 $\pm$ 5.59} \\
 &  & RepNodes  & 25.05 $\pm$ 0.80 & \cellcolor{green!30}\textbf{66.43 $\pm$ 3.35} & \cellcolor{green!30}{74.28 $\pm$ 1.06} & 84.22 $\pm$ 1.63 & 45.56 $\pm$ 2.48 & 86.54 $\pm$ 0.57 & \cellcolor{green!30}\textbf{50.69 $\pm$ 4.84} & 40.00 $\pm$ 6.09 & \cellcolor{green!30}{59.20 $\pm$ 5.22} \\
 &  & RepEdges  & 25.05 $\pm$ 0.80 & \cellcolor{green!30}\textbf{66.43 $\pm$ 3.35} & \cellcolor{green!30}\textbf{74.58 $\pm$ 1.22} & \cellcolor{green!30}\textbf{85.04 $\pm$ 0.81} & \cellcolor{green!30}\textbf{51.11 $\pm$ 4.65} & \cellcolor{green!30}\textbf{86.76 $\pm$ 0.22} & \cellcolor{green!30}{45.23 $\pm$ 3.92} & 46.67 $\pm$ 8.43 & \cellcolor{green!30}\textbf{61.60 $\pm$ 7.27} \\
\cline{2-12}
 & \multirow[t]{4}{*}{No} & Baseline  & 24.97 $\pm$ 0.32 & 66.34 $\pm$ 2.03 & \textbf{70.12 $\pm$ 0.55} & \textbf{82.67 $\pm$ 0.67} & \textbf{46.67 $\pm$ 4.68} & \textbf{82.46 $\pm$ 0.59} & 47.15 $\pm$ 2.24 & \textbf{47.78 $\pm$ 2.87} & \textbf{55.20 $\pm$ 1.69} \\
 &  & MN  & \underline{25.29 $\pm$ 0.58} & \underline{67.31 $\pm$ 2.11} & \multicolumn{1}{c}{OOM} & 81.56 $\pm$ 0.85 & 44.44 $\pm$ 3.60 & \multicolumn{1}{c}{OOM} & 40.61 $\pm$ 6.70 & 36.67 $\pm$ 6.84 & 53.60 $\pm$ 6.67 \\
 &  & RepNodes  & \cellcolor{green!30}\textbf{26.00 $\pm$ 0.59} & \cellcolor{green!30}\textbf{68.81 $\pm$ 0.80} & 68.92 $\pm$ 0.47 & 82.07 $\pm$ 1.27 & 45.56 $\pm$ 2.34 & 81.56 $\pm$ 0.62 & \cellcolor{green!30}\textbf{56.50 $\pm$ 2.45} & 41.11 $\pm$ 10.21 & 53.60 $\pm$ 6.85 \\
 &  & RepEdges  & \cellcolor{green!30}\textbf{26.00 $\pm$ 0.59} & \cellcolor{green!30}\textbf{68.81 $\pm$ 0.80} & 68.55 $\pm$ 1.52 & 82.37 $\pm$ 1.41 & 46.67 $\pm$ 2.87 & 81.54 $\pm$ 0.60 & \cellcolor{green!30}{55.79 $\pm$ 7.81} & 44.44 $\pm$ 9.07 & 55.20 $\pm$ 4.13 \\
\cline{1-12} \cline{2-12}
\midrule
\multirow[t]{8}{*}{GIN} & \multirow[t]{4}{*}{Yes} & Baseline  & \textbf{24.68 $\pm$ 0.18} & 67.58 $\pm$ 3.08 & 73.13 $\pm$ 1.38 & 83.04 $\pm$ 1.03 & 51.11 $\pm$ 4.65 & \textbf{84.89 $\pm$ 0.44} & 65.23 $\pm$ 2.04 & 58.89 $\pm$ 3.04 & 56.00 $\pm$ 2.83 \\
 &  & MN  & 23.46 $\pm$ 1.61 & 62.95 $\pm$ 2.21 & \multicolumn{1}{c}{OOM} & 82.74 $\pm$ 1.12 & 43.33 $\pm$ 2.34 & \multicolumn{1}{c}{OOM} & \underline{65.65 $\pm$ 3.58} & 56.67 $\pm$ 4.38 & 51.20 $\pm$ 4.13 \\
 &  & RepNodes  & 24.61 $\pm$ 0.00 & 65.90 $\pm$ 1.91 & \cellcolor{green!30}\textbf{75.00 $\pm$ 0.70} & \cellcolor{green!30}\textbf{83.93 $\pm$ 0.85} & \cellcolor{green!30}\textbf{54.44 $\pm$ 7.24} & 84.73 $\pm$ 0.53 & \cellcolor{green!30}\textbf{66.81 $\pm$ 3.30} & \cellcolor{green!30}\textbf{61.11 $\pm$ 6.80} & 53.60 $\pm$ 4.56 \\
 &  & RepEdges  & 24.61 $\pm$ 0.00 & \cellcolor{green!30}\textbf{68.63 $\pm$ 2.84} & \cellcolor{green!30}\textbf{75.00 $\pm$ 0.70} & \cellcolor{green!30}\textbf{83.93 $\pm$ 0.85} & \cellcolor{green!30}\textbf{54.44 $\pm$ 7.24} & 84.77 $\pm$ 0.65 & \cellcolor{green!30}{65.65 $\pm$ 3.80} & \cellcolor{green!30}\textbf{61.11 $\pm$ 6.80} & \cellcolor{green!30}\textbf{61.60 $\pm$ 4.56} \\
\cline{2-12}
 & \multirow[t]{4}{*}{No} & Baseline  & 24.78 $\pm$ 0.38 & 63.35 $\pm$ 0.68 & \textbf{68.67 $\pm$ 0.72} & 82.89 $\pm$ 0.80 & 53.33 $\pm$ 8.76 & \textbf{82.16 $\pm$ 0.68} & 64.65 $\pm$ 4.07 & 54.44 $\pm$ 2.34 & \textbf{58.00 $\pm$ 4.32} \\
 &  & MN  & 24.45 $\pm$ 0.99 & 61.94 $\pm$ 4.53 & \multicolumn{1}{c}{OOM} & 82.59 $\pm$ 0.80 & 45.56 $\pm$ 4.27 & \multicolumn{1}{c}{OOM} & \underline{64.73 $\pm$ 3.47} & \underline{\textbf{58.89 $\pm$ 2.79}} & 49.20 $\pm$ 3.69 \\
 &  & RepNodes  & 24.61 $\pm$ 0.00 & \cellcolor{green!30}\textbf{71.89 $\pm$ 2.76} & 68.55 $\pm$ 1.51 & \cellcolor{green!30}\textbf{82.96 $\pm$ 0.65} & \cellcolor{green!30}{53.33 $\pm$ 7.94} & 39.57 $\pm$ 0.30 & \cellcolor{green!30}\textbf{67.27 $\pm$ 0.94} & \cellcolor{green!30}\textbf{58.89 $\pm$ 2.87} & 54.40 $\pm$ 8.68 \\
 &  & RepEdges  & \cellcolor{green!30}\textbf{24.88 $\pm$ 0.48} & \cellcolor{green!30}\textbf{71.89 $\pm$ 2.76} & 68.55 $\pm$ 1.51 & \cellcolor{green!30}\textbf{82.96 $\pm$ 0.65} & \cellcolor{green!30}\textbf{56.67 $\pm$ 4.38} & 39.44 $\pm$ 0.43 & \cellcolor{green!30}{65.38 $\pm$ 2.29} & \cellcolor{green!30}\textbf{58.89 $\pm$ 2.87} & 57.60 $\pm$ 2.07 \\
\cline{1-12} \cline{2-12}
\bottomrule
\end{tabular}}
\caption{Node classification accuracies at $L=L^*$ (best baseline $8\geq L>2$). In green are highlighted the cases in which RAwR is better than the baseline. We highlight with underlines cases where MN is better than the baseline and the overall best performance in bold. For RepNodes and RepEdges, the results refer to the $\varepsilon$ selected with the grid search on the validation set.}
\label{tab:ncq3}
\end{table*}

\begin{table*}[h]
\centering
\resizebox{\textwidth}{!}{%
\begin{tabular}{llllllllllll}
\toprule
 &  & Dataset & Actor & Chameleon & Citeseer & Cora & Cornell & PubMed & Squirrel & Texas & Wisconsin \\
Model & Features & Method &  &  &  &  &  &  &  &  &  \\
\midrule
\multirow[t]{8}{*}{GCN} & \multirow[t]{4}{*}{Yes} & Baseline  & 25.74 $\pm$ 0.93 & 65.02 $\pm$ 0.59 & 75.66 $\pm$ 0.65 & 84.07 $\pm$ 0.94 & 48.89 $\pm$ 2.48 & 87.63 $\pm$ 0.24 & 61.65 $\pm$ 0.22 &\textbf{44.44 $\pm$ 3.93} & 58.40 $\pm$ 3.58 \\
 &  & MN  & \underline{25.97 $\pm$ 0.88} & \underline{66.30 $\pm$ 0.84} & \multicolumn{1}{c}{OOM} & \underline{84.96 $\pm$ 0.63} & 47.78 $\pm$ 2.87 & \multicolumn{1}{c}{OOM} & 61.17 $\pm$ 0.29 & 42.22 $\pm$ 2.87 & \underline{61.60 $\pm$ 7.35} \\
 &  & RepNodes  & \cellcolor{green!30}\textbf{26.45 $\pm$ 0.71} & \cellcolor{green!30}\textbf{73.83 $\pm$ 0.50} & \cellcolor{green!30}\textbf{76.02 $\pm$ 1.08} & \cellcolor{green!30}\textbf{85.26 $\pm$ 0.61} & \cellcolor{green!30}\textbf{56.67 $\pm$ 2.48} & \cellcolor{green!30}\textbf{87.77 $\pm$ 0.21} & \cellcolor{green!30}\textbf{65.77 $\pm$ 0.38} & 43.33 $\pm$ 2.48 & \cellcolor{green!30}\textbf{65.60 $\pm$ 2.19} \\
 &  & RepEdges  & \cellcolor{green!30}{26.21 $\pm$ 0.80} & \cellcolor{green!30}{71.37 $\pm$ 0.31} & \cellcolor{green!30}\textbf{76.02 $\pm$ 1.20} & \cellcolor{green!30}{84.96 $\pm$ 0.77} & \cellcolor{green!30}{52.22 $\pm$ 4.97} & \cellcolor{green!30}{87.75 $\pm$ 0.19} & \cellcolor{green!30}{63.46 $\pm$ 0.19} & 43.33 $\pm$ 10.69 & \cellcolor{green!30}{64.00 $\pm$ 2.83} \\
\cline{2-12}
 & \multirow[t]{4}{*}{No} & Baseline  & \textbf{25.45 $\pm$ 0.78} & 67.84 $\pm$ 0.93 & 69.73 $\pm$ 0.57 & \textbf{83.48 $\pm$ 0.19} & 46.67 $\pm$ 2.87 & \textbf{82.12 $\pm$ 0.06} & 60.85 $\pm$ 0.26 & \textbf{47.78 $\pm$ 7.94} & 54.40 $\pm$ 3.37 \\
 &  & MN  & 25.39 $\pm$ 0.90 & \underline{68.24 $\pm$ 0.35} & \multicolumn{1}{c}{OOM} & 83.24 $\pm$ 0.32 & \underline{48.89 $\pm$ 4.27} & \multicolumn{1}{c}{OOM} & 60.73 $\pm$ 0.53 & 41.11 $\pm$ 2.79 & \underline{62.40 $\pm$ 5.57} \\
 &  & RepNodes  & 25.39 $\pm$ 1.08 & \cellcolor{green!30}\textbf{75.07 $\pm$ 0.23} & \cellcolor{green!30}\textbf{70.84 $\pm$ 0.13} & 83.33 $\pm$ 0.00 & \cellcolor{green!30}\textbf{55.56 $\pm$ 0.00} & 81.95 $\pm$ 0.19 & \cellcolor{green!30}\textbf{65.42 $\pm$ 0.76} & 43.33 $\pm$ 2.34 & \cellcolor{green!30}\textbf{68.00 $\pm$ 0.00} \\
 &  & RepEdges  & 25.39 $\pm$ 1.08 & \cellcolor{green!30}{71.72 $\pm$ 0.28} & \cellcolor{green!30}{69.79 $\pm$ 0.29} & 83.41 $\pm$ 0.16 & \cellcolor{green!30}{51.11 $\pm$ 2.34} & 81.93 $\pm$ 0.23 & \cellcolor{green!30}{63.50 $\pm$ 0.72} & 43.33 $\pm$ 10.08 & \cellcolor{green!30}{66.40 $\pm$ 2.07} \\
\cline{1-12} \cline{2-12}
\midrule
\multirow[t]{8}{*}{GAT} & \multirow[t]{4}{*}{Yes} & Baseline  & 25.39 $\pm$ 1.02 & 60.09 $\pm$ 5.47 & 74.04 $\pm$ 0.39 & 84.89 $\pm$ 0.61 & 46.67 $\pm$ 3.04 & 86.67 $\pm$ 0.35 & 43.50 $\pm$ 1.55 & \textbf{48.89 $\pm$ 2.48} & 55.20 $\pm$ 5.22 \\
 &  & MN  & \underline{\textbf{26.47 $\pm$ 0.57}} & 60.00 $\pm$ 5.20 & \multicolumn{1}{c}{OOM} & 84.44 $\pm$ 0.72 & 44.44 $\pm$ 0.00 & \multicolumn{1}{c}{OOM} & \underline{45.77 $\pm$ 0.88} & 43.33 $\pm$ 8.61 & \underline{56.80 $\pm$ 5.59} \\
 &  & RepNodes  & \cellcolor{green!30}\textbf{26.47 $\pm$ 0.61} & \cellcolor{green!30}\textbf{68.28 $\pm$ 3.70} & \cellcolor{green!30}{74.52 $\pm$ 1.32} & 84.52 $\pm$ 0.80 & 46.67 $\pm$ 3.04 & 86.54 $\pm$ 0.57 & \cellcolor{green!30}\textbf{50.69 $\pm$ 4.84} & 43.33 $\pm$ 9.13 & \cellcolor{green!30}{59.20 $\pm$ 5.22} \\
 &  & RepEdges  & \cellcolor{green!30}\textbf{26.47 $\pm$ 0.61} & \cellcolor{green!30}{67.84 $\pm$ 4.28} & \cellcolor{green!30}\textbf{74.58 $\pm$ 1.22} & \cellcolor{green!30}\textbf{85.04 $\pm$ 0.81} & \cellcolor{green!30}\textbf{51.11 $\pm$ 4.65} & \cellcolor{green!30}\textbf{86.76 $\pm$ 0.22} & \cellcolor{green!30}{46.77 $\pm$ 8.01} & 46.67 $\pm$ 8.43 & \cellcolor{green!30}\textbf{61.60 $\pm$ 7.27} \\
\cline{2-12}
 & \multirow[t]{4}{*}{No} & Baseline  & 24.97 $\pm$ 0.32 & 66.34 $\pm$ 2.03 & \textbf{70.12 $\pm$ 0.55} & \textbf{82.67 $\pm$ 0.67} & 46.67 $\pm$ 4.68 & \textbf{82.46 $\pm$ 0.59} & 47.15 $\pm$ 2.24 & \textbf{47.78 $\pm$ 2.87} & 55.20 $\pm$ 1.69 \\
 &  & MN  & \underline{25.29 $\pm$ 0.58} & \underline{67.31 $\pm$ 2.11} & \multicolumn{1}{c}{OOM} & 81.56 $\pm$ 0.85 & 44.44 $\pm$ 0.00 & \multicolumn{1}{c}{OOM} & \underline{47.62 $\pm$ 2.87} & 37.78 $\pm$ 6.65 & \underline{59.80 $\pm$ 2.75} \\
 &  & RepNodes  & \cellcolor{green!30}\textbf{26.00 $\pm$ 0.59} & \cellcolor{green!30}{68.81 $\pm$ 0.80} & 68.92 $\pm$ 0.47 & 82.07 $\pm$ 1.27 & \cellcolor{green!30}\textbf{48.89 $\pm$ 2.34} & 82.09 $\pm$ 0.98 & \cellcolor{green!30}\textbf{56.50 $\pm$ 2.45} & 42.78 $\pm$ 9.46 & \cellcolor{green!30}{59.60 $\pm$ 2.95} \\
 &  & RepEdges  & \cellcolor{green!30}\textbf{26.00 $\pm$ 0.59} & \cellcolor{green!30}\textbf{69.34 $\pm$ 2.26} & 68.55 $\pm$ 1.52 & 82.37 $\pm$ 1.41 & \cellcolor{green!30}{47.78 $\pm$ 4.68} & 82.08 $\pm$ 0.97 & \cellcolor{green!30}{55.79 $\pm$ 7.81} & 44.44 $\pm$ 9.07 & \cellcolor{green!30}\textbf{63.20 $\pm$ 4.13} \\
\cline{1-12} \cline{2-12}
\midrule
\multirow[t]{8}{*}{GIN} & \multirow[t]{4}{*}{Yes} & Baseline  & 24.68 $\pm$ 0.18 & 67.58 $\pm$ 3.08 & 73.13 $\pm$ 1.38 & 83.04 $\pm$ 1.03 & 51.11 $\pm$ 4.65 & \textbf{84.89 $\pm$ 0.44} & 65.23 $\pm$ 2.04 & 58.89 $\pm$ 3.04 & 56.00 $\pm$ 2.83 \\
 &  & MN  & \underline{24.92 $\pm$ 0.69} & 62.95 $\pm$ 2.21 & \multicolumn{1}{c}{OOM} & 82.74 $\pm$ 1.12 & 45.00 $\pm$ 4.10 & \multicolumn{1}{c}{OOM} & \underline{65.65 $\pm$ 3.58} & 56.67 $\pm$ 4.38 & 54.80 $\pm$ 3.79 \\
 &  & RepNodes  & \cellcolor{green!30}\textbf{25.66 $\pm$ 1.74} & \cellcolor{green!30}\textbf{71.28 $\pm$ 1.57} & \cellcolor{green!30}\textbf{75.00 $\pm$ 0.70} & \cellcolor{green!30}\textbf{83.93 $\pm$ 0.85} & \cellcolor{green!30}{55.56 $\pm$ 11.79} & 84.73 $\pm$ 0.53 & \cellcolor{green!30}\textbf{66.81 $\pm$ 3.30} & \cellcolor{green!30}\textbf{61.11 $\pm$ 6.80} & \cellcolor{green!30}{60.80 $\pm$ 1.79} \\
 &  & RepEdges  & \cellcolor{green!30}\textbf{25.66 $\pm$ 1.74} & \cellcolor{green!30}\textbf{71.28 $\pm$ 1.57} & \cellcolor{green!30}\textbf{75.00 $\pm$ 0.70} & \cellcolor{green!30}\textbf{83.93 $\pm$ 0.85} & \cellcolor{green!30}\textbf{57.78 $\pm$ 3.04} & 84.77 $\pm$ 0.65 & \cellcolor{green!30}{65.65 $\pm$ 3.80} & \cellcolor{green!30}{61.11 $\pm$ 6.80} & \cellcolor{green!30}\textbf{62.40 $\pm$ 4.56} \\
\cline{2-12}
 & \multirow[t]{4}{*}{No} & Baseline  & 24.78 $\pm$ 0.38 & 63.35 $\pm$ 0.68 & \textbf{68.67 $\pm$ 0.72} & 82.89 $\pm$ 0.80 & 53.33 $\pm$ 8.76 & \textbf{82.16 $\pm$ 0.68} & 64.65 $\pm$ 4.07 & 54.44 $\pm$ 2.34 & 58.00 $\pm$ 4.32 \\
 &  & MN  & \underline{\textbf{25.13 $\pm$ 0.91}} & 62.93 $\pm$ 4.46 & \multicolumn{1}{c}{OOM} & 82.59 $\pm$ 0.80 & 45.56 $\pm$ 2.28 & \multicolumn{1}{c}{OOM} & \underline{64.73 $\pm$ 3.47} & \underline{\textbf{58.89 $\pm$ 2.79}} & 56.60 $\pm$ 3.95 \\
 &  & RepNodes  & \cellcolor{green!30}\textbf{25.13 $\pm$ 0.93} & \cellcolor{green!30}\textbf{71.89 $\pm$ 2.76} & 68.55 $\pm$ 1.51 & \cellcolor{green!30}\textbf{82.96 $\pm$ 0.65} & \cellcolor{green!30}{55.56 $\pm$ 0.00} & 81.72 $\pm$ 0.80 & \cellcolor{green!30}\textbf{67.27 $\pm$ 0.94} & \cellcolor{green!30}\textbf{58.89 $\pm$ 2.87} & \cellcolor{green!30}\textbf{58.40 $\pm$ 3.37} \\
 &  & RepEdges  & \cellcolor{green!30}\textbf{25.13 $\pm$ 0.93} & \cellcolor{green!30}\textbf{71.89 $\pm$ 2.76} & 68.55 $\pm$ 1.51 & \cellcolor{green!30}\textbf{82.96 $\pm$ 0.65} & \cellcolor{green!30}\textbf{56.67 $\pm$ 4.38} & 81.31 $\pm$ 0.61 & \cellcolor{green!30}{65.38 $\pm$ 2.29} & \cellcolor{green!30}\textbf{58.89 $\pm$ 2.87} & \cellcolor{green!30}\textbf{58.40 $\pm$ 6.31} \\
\cline{1-12} \cline{2-12}
\bottomrule
\end{tabular}}
\caption{Best Accuracy Baseline at $8\geq L\geq2$ vs Best Accuracy Augmentations at $8\geq L\geq2$ for Node Classification. In green are highlighted the cases in which RAwR is better than the baseline. We highlight with underlines cases where MN is better than the baseline and mark the best perfromance in bold. For RepNodes and RepEdges, the results refer to the $\varepsilon$ selected with the grid search on the validation set.}
\label{tab:ncq4}
\end{table*}

\begin{table*}[h]
\centering
\scalebox{0.7}{
\begin{tabular}{lccccccccc}
\toprule
 & Actor & Chameleon & Citeseer & Cora & Cornell & PubMed & Squirrel & Texas & Wisconsin \\
\midrule
GCN & -0.18 & \cellcolor{green!20}\textbf{+11.54} & \cellcolor{green!20}\textbf{+3.61} & \cellcolor{green!20}\textbf{+1.00} & \textbf{-2.22} & \cellcolor{green!20}+0.51 & \cellcolor{green!20}\textbf{+7.08} & \cellcolor{green!20}\textbf{+10.00} & \cellcolor{green!20}\textbf{+1.60} \\
GAT & \cellcolor{green!20}+0.08 & \cellcolor{green!20}\textbf{+2.47} & \cellcolor{green!20}+0.66 & \cellcolor{green!20}+0.15 & +0.00 & \cellcolor{green!20}+0.90 & \cellcolor{green!20}+0.88 & +0.00 & \cellcolor{green!20}\textbf{+2.40} \\
GIN & \cellcolor{green!20}+0.26 & \cellcolor{green!20}\textbf{+8.63} & \cellcolor{green!20}\textbf{+2.95} & -0.59 & \cellcolor{green!20}\textbf{+7.78} & \cellcolor{green!20}+0.90 & \cellcolor{green!20}\textbf{+2.46} & \cellcolor{green!20}\textbf{+18.89} & \textbf{-1.60} \\
\bottomrule
\end{tabular}
}
\caption{Difference in mean test accuracy between RAwR RepNodes and a control variant where the partition is replaced by a random assignment. We highlight in green positive differences and we report in bold values greater than $1$ in absolute value.}
\label{tab:approx_vs_random_best_delta_sameeps_gat_gin}
\end{table*}

\begin{table}[H]
\centering
\resizebox{\textwidth}{!}{%
\begin{tabular}{llcccccccccc}
\hline
Features & Method & Actor & Chameleon & Citeseer & Cora & Cornell & PubMed & Squirrel & Texas & Wisconsin & Avg.Rank \\
\hline
Yes & GCN+RepNodes & 25.11 $\pm$ 0.82 & \textbf{56.25 $\pm$ 12.18} & \textbf{75.24 $\pm$ 0.39} & 80.15 $\pm$ 4.43 & 46.67 $\pm$ 3.86 & \textbf{85.25 $\pm$ 1.34} & \textbf{45.12 $\pm$ 11.32} & 39.21 $\pm$ 4.03 & 57.03 $\pm$ 7.15 & \textbf{2.83} \\
 & GCN+RepEdges & 25.06 $\pm$ 0.73 & \underline{55.08 $\pm$ 11.95} & \textbf{75.24 $\pm$ 0.39} & \textbf{81.27 $\pm$ 3.62} & 47.78 $\pm$ 4.90 & \underline{85.25 $\pm$ 1.37} & \underline{44.94 $\pm$ 10.50} & 39.21 $\pm$ 4.03 & 55.09 $\pm$ 6.66 & 3.00 \\
 & GRAIN & 29.13 $\pm$ 1.29 & 44.32 $\pm$ 1.38 & \underline{67.41 $\pm$ 2.64} & \underline{80.44 $\pm$ 3.18} & 63.33 $\pm$ 11.52 & 80.05 $\pm$ 0.81 & 32.77 $\pm$ 1.42 & 71.11 $\pm$ 13.26 & \underline{68.80 $\pm$ 11.45} & 3.11 \\
 & GRAIN+RepNodes & \underline{29.42 $\pm$ 1.59} & 46.08 $\pm$ 1.45 & 65.84 $\pm$ 2.38 & 80.00 $\pm$ 2.36 & \textbf{68.89 $\pm$ 12.17} & 79.76 $\pm$ 1.10 & 32.58 $\pm$ 0.57 & \underline{73.33 $\pm$ 4.65} & \textbf{75.20 $\pm$ 10.35} & 3.17 \\
 & GRAIN+RepEdges & \textbf{29.47 $\pm$ 2.20} & 45.73 $\pm$ 1.80 & 66.02 $\pm$ 3.18 & 80.15 $\pm$ 2.61 & \underline{67.78 $\pm$ 10.69} & 79.82 $\pm$ 0.89 & 32.54 $\pm$ 1.40 & \textbf{76.67 $\pm$ 4.65} & \textbf{75.20 $\pm$ 11.45} & \underline{2.89} \\
\hline
No & GCN+RepNodes & 24.89 $\pm$ 0.66 & \textbf{55.13 $\pm$ 13.64} & \textbf{57.65 $\pm$ 15.80} & 69.39 $\pm$ 12.90 & \textbf{48.10 $\pm$ 4.62} & 59.07 $\pm$ 19.73 & \textbf{45.97 $\pm$ 10.95} & 39.37 $\pm$ 3.87 & \underline{57.09 $\pm$ 6.49} & \underline{2.72} \\
 & GCN+RepEdges & 24.77 $\pm$ 0.54 & \underline{54.97 $\pm$ 12.59} & \underline{56.29 $\pm$ 19.14} & \textbf{73.61 $\pm$ 11.36} & \underline{47.62 $\pm$ 4.08} & 59.08 $\pm$ 19.63 & \underline{45.47 $\pm$ 10.35} & 39.37 $\pm$ 3.87 & 56.23 $\pm$ 7.01 & 3.06 \\
 & GRAIN & 26.55 $\pm$ 1.03 & 49.96 $\pm$ 2.50 & 44.94 $\pm$ 3.05 & 70.67 $\pm$ 3.88 & 45.56 $\pm$ 9.13 & \textbf{65.01 $\pm$ 5.03} & 30.38 $\pm$ 0.67 & 46.67 $\pm$ 16.48 & 56.80 $\pm$ 10.35 & 3.78 \\
 & GRAIN+RepNodes & \textbf{26.84 $\pm$ 0.95} & 51.28 $\pm$ 0.97 & 46.45 $\pm$ 1.74 & \underline{70.89 $\pm$ 3.17} & 45.56 $\pm$ 12.04 & 64.73 $\pm$ 2.81 & 31.35 $\pm$ 1.25 & \textbf{52.22 $\pm$ 13.38} & \textbf{57.60 $\pm$ 10.81} & \textbf{2.44} \\
 & GRAIN+RepEdges & \underline{26.82 $\pm$ 0.72} & 51.10 $\pm$ 1.87 & 46.39 $\pm$ 1.76 & \underline{70.89 $\pm$ 3.17} & 46.67 $\pm$ 10.83 & \underline{64.83 $\pm$ 3.12} & 31.27 $\pm$ 1.95 & \underline{50.00 $\pm$ 16.20} & 56.80 $\pm$ 10.35 & 3.00 \\
\hline
\end{tabular}}
\caption{Comparison among GCN+RAwR and GRAIN-based methods. For each RAwR variant, the best epsilon per dataset is selected with a grid search on the validation set. Best is in bold; second-best is underlined. Avg.Rank uses lower-is-better with the same styling.}
\label{tab:gcn_grain_rawr_comparison}
\end{table}

\subsection{Machines adopted}
Experiments on SRL, RAwR and Master Node have been run on a
machine equipped with an Intel(R) Xeon(R) CPU E7-4830
with 500 GB RAM, while the other experiments have been performed on a Lenovo ThinkStation PGX.

\newpage

\end{document}